\documentclass{article}

\usepackage{arxiv}

\usepackage[utf8]{inputenc} 
\usepackage[T1]{fontenc}    
\usepackage{hyperref}       
\usepackage{url}            
\usepackage{booktabs}       
\usepackage{amsfonts}       
\usepackage{nicefrac}       
\usepackage{microtype}      
\usepackage{lipsum}
\usepackage{ctable}
\usepackage{wrapfig}
\usepackage{subcaption}
\usepackage{graphicx}
\usepackage{amsmath}
\usepackage{amssymb}
\usepackage{cleveref}
\usepackage{multirow}

\title{Deep Learning for Post-Processing Ensemble Weather Forecasts}

\author{%
  Peter Gr\"{o}nquist \\
  ETH Z\"{u}rich \\
  \texttt{petergro@student.ethz.ch}
  \And
  Chengyuan Yao \\
  ETH Z\"{u}rich \\
  \texttt{chyao@student.ethz.ch}
  \And
  Tal Ben-Nun \\
  ETH Z\"{u}rich \\
  \texttt{tal.bennun@inf.ethz.ch}
  \And
  Nikoli Dryden \\
  ETH Z\"{u}rich \\
  \texttt{nikoli.dryden@inf.ethz.ch}
  \And
  Peter Dueben \\
  ECMWF \\
  \texttt{peter.dueben@ecmwf.int}
  \And
  Shigang Li \\
  ETH Z\"{u}rich \\
  \texttt{shigang.li@inf.ethz.ch}
  \And
  Torsten Hoefler \\
  ETH Z\"{u}rich \\
  \texttt{htor@inf.ethz.ch}
}

\begin{document}
\maketitle

\begin{abstract}
Quantifying uncertainty in weather forecasts is critical, especially for predicting extreme weather events. This is typically accomplished with ensemble prediction systems, which consist of many perturbed numerical weather simulations, or trajectories, run in parallel. These systems are associated with a high computational cost and often involve statistical post-processing steps to inexpensively improve their raw prediction qualities. We propose a mixed model that uses only a subset of the original weather trajectories combined with a post-processing step using deep neural networks. These enable the model to account for non-linear relationships that are not captured by current numerical models or post-processing methods.
Applied to global data, our mixed models achieve a relative improvement in ensemble forecast skill (CRPS) of over 14\%. Furthermore, we demonstrate that the improvement is larger for extreme weather events on select case studies. We also show that our post-processing can use fewer trajectories to achieve comparable results to the full ensemble. By using fewer trajectories, the computational costs of an ensemble prediction system can be reduced, allowing it to run at higher resolution and produce more accurate forecasts.
\end{abstract}


\section{Introduction}
Operational weather predictions have a large impact on society. They influence individuals on a daily basis, and in more severe cases, save lives and property by predicting extreme events such as tropical cyclones. However, developing reliable weather prediction systems is a difficult task due to the complexity of the Earth System and the chaotic behaviour of its components. Small errors introduced by observations, their assimilation, and the forecast model configuration escalate chaotically, leading to a significant loss in forecast skill within a week. 
Numerical Weather Prediction (NWP) is based on computer models solving complex partial differential equations at limited resolution. To be useful, weather forecasts try to estimate the uncertainties in predictions using ensemble simulations, where a forecast model is run a number of times from slightly different initial conditions, parameter values, and stochastic forcing. The resulting spread of predictions among ensemble members provides an estimate for the prediction uncertainty. This enables us to estimate the probability of, for example, precipitation for a specific location and time of day as well as the probability of a tropical cyclone hitting a large city.


In this paper we will focus on post-processing ensemble predictions performed at the European Centre for Medium-Range Weather Forecasts (ECMWF)\cite{ecmwfensembles} using deep neural networks (DNNs). ECMWF runs an operational forecast that consists of one high resolution (9 km grid) deterministic forecast (HRES), and an ensemble (ENS) with 51 members at a lower resolution (18 km), of which one is the unperturbed control trajectory. Each ensemble member starts from slightly different initial conditions and uses a different stochastic forcing in the physical parameterisation schemes of subgrid-scale processes --- so-called stochastic parameterisation schemes.
While ensemble methods have become a standard tool for numerical weather predictions, there is an ongoing discussion on how many ensemble members should be used. Larger ensembles allow for a better sampling of the probability density function (PDF) of predictions. 
However, computing power is limited and forecasts are bound by strict operational time windows of a couple of hours. Smaller ensembles would therefore allow individual members to run at higher resolution, likely resulting in better forecasts by each ensemble member~\cite{ecmwfforcastincrease}. 


The demand for ever more precise and dependable forecasts has led NWP methods to rank amongst the scientific domains with the most significant demand for supercomputing time~\cite{schulthess-exascale-climats,Neumann2019,crclim,Dueben2020}. 
As such, the NWP field is constantly looking for new methods to improve accuracy and reduce the computational cost of its models. 
This is where the recent advances in DNNs~\cite{lecun2015deep} become relevant. The breadth of tasks and efficient inference DNNs enable has made them a very attractive option for improving weather forecasts~\cite{doi:10.1029/2018GL078510,rasp18ens,rasp2018deep,gmd-11-3999-2018,rasp2020weatherbench}. Related studies have also shown their capabilities for predicting chaotic behavior~\cite{chaos,petros}. 
However, the full potential of these methods remains unexplored in many areas of NWP.


We use convolutional neural networks (CNNs)~\cite{LeCun1999} and locally connected networks (LCNs)~\cite{coates13} to both improve forecast skill and reduce the computational requirements for NWP. 
We approach these goals through three different tasks: Uncertainty Quantification, Bias Correction, and PDF Calibration.
Each is a different task that is addressed with a different neural network. First, uncertainty quantification is usually performed by examining the spread (standard deviation) of the forecasting ensemble. Here, we train a neural network to produce a similar spread as an ensemble, using only a small fraction of the ensemble members as input. We achieve a relative RMSE improvement of over 16\% in forecast ability compared to using five of ten ensemble members to predict temperature. Second, we train a neural network to predict a point-wise bias to account for local trends in weather patterns. This results in a relative RMSE improvement of 7.9\% on temperature. Lastly, we calibrate the ensemble PDF given a bias-corrected input, using our uncertainty quantification network. This results in a forecast skill increase of over 14.5\% using only half the trajectories of a full ensemble.
The reduced number of input ensemble forecasts also allows NWP to be run at a fraction of the cost of additional trajectories. Prediction time is further reduced by making use of high throughput graphics processing units (GPUs) for DNN inference.

Our code and data are publicly available\footnote{\url{https://github.com/spcl/deep-weather}}.

\subsection{Related Work}
There have been many works leveraging the modelling capabilities of neural networks (NNs) for NWP. Early attempts at applying shallow NNs showed success in emulating physical processes and saving computational power~\cite{doi:10.1175}. Since then, building on recent DNN developments, much effort has gone into applying NNs to weather nowcasting\cite{xingjian2015convolutional,shi2017deep,heye2017precipitation,agrawal2019machine}. 
Nowcasting focuses on the emulation of physical processes for short term (up to six hours), high-resolution forecasts. Other works have also shown the significant capabilities of DNNs to predict longer ranging forecasts and extreme weather patterns\cite{kurth2018exascale,larraondo2019datadriven,doi:10.1029/2019MS001705,doi:10.1175/WAF-D-18-0183.1,liu2016application,rasp2020weatherbench}.

In contrast, we focus on the post-processing of operational medium-range ensemble forecasts and the prediction of extreme weather events.  
Post-processing ensemble outputs has been a long-standing effort in the weather forecasting community. Methods such as Ensemble Model Output Statistics (EMOS)\cite{doi:10.1175/MWR2904.1} and Bayesian Model Averaging (BMA)\cite{doi:10.1175/MWR2906.1} currently allow for improvements of the raw ensemble forecast skill. 
Hamill and Whitaker~\cite{doi:10.1175/MWR3468.1} show initial explorations of those techniques on re-forecast datasets, also used in this paper, for temperature at 850 hPa (T850) and geopotential at 500 hPa (Z500). 
Advances in neural networks have only recently reached the field of ensemble models in weather forecasting, focusing on its application to specific weather stations~\cite{rasp18ens,baran2020machine} or global interpolations~\cite{Wunc}. 
We expand on this work by applying DNNs on the novel task of improving the forecast skill for global predictions, specifically extreme weather forecasts, while reducing their computational costs.

\section{Data}
\label{sec:data}
The quality of ensemble forecasts has improved significantly over the last decades and ensemble predictions are using increased resolutions and numbers of trajectories. 
Learning from past ensemble predictions would therefore lead to inconsistencies as the correction for mean and spread would need to adjust for changes in the quality of predictions over time.
To address this, \emph{re-forecasts}~\cite{doi:10.1175/BAMS-87-1-33} apply current state-of-the-art forecast models to past measurements.

We use data from re-forecast ensemble experiments at ECMWF. These are routinely generated to provide an estimate of the "climate" of the forecast model for each date of the year, which can be used to remove model drifts during post-processing and for measuring the generic skill of the forecast system \cite{vitart2014,vitart2019}. 
The re-forecast experiments run a 10-member ensemble (ENS10) and an unperturbed control experiment. Simulations use 91 vertical levels, the spectral representation of the model fields is truncated at the global wavenumber 639, and a cubic octahedral reduced Gaussian grid is used for the representation of model fields in grid-point space that provides an approximately uniform distribution of grid-cells on the sphere with a 18 km grid-spacing (the "TCo639" grid~\cite{Wedi2014}). Simulations are performed with the same system for 1999-2017, with two forecast simulations starting each week, providing a large dataset with consistent forecast quality.

To fully train our networks and evaluate their forecast skill we also need ground truth weather conditions at specific forecast lead times. For this, we use the fifth major ECMWF ReAnalysis (ERA5)\cite{era5}, which includes data on weather from 1979 up to the present\footnote{Available for download under \url{https://cds.climate.copernicus.eu/}}. Compared to re-forecasts, reanalysis datasets are produced by applying a constant stream of observations through state-of-the-art data assimilation on state-of-the-art forecast models used for decade-long simulations. 
In ERA5, this process generates reanalysis fields that are available at an hourly frequency for over 300 parameters.

\subsection{Data Selection}
\label{subsec:data-selection}
We use the ENS10 dataset for our forecasts and make use of ERA5's constant data assimilation product as ground truth, given the ENS10 forecast lead times. 
ENS10 and ERA5 both provide global data, which we interpolate to a latitude/longitude grid with a 0.5 degree resolution. We do this to avoid the native grid that was used within simulations, as it is unstructured in the longitudinal direction. 
While the use of latitude/longitude grids does lead to over-saturation of gridpoints in the poles, it simplifies our models; we leave the use of unstructured grids to future research. 
We also focus on a single pressure level for each model. When predicting temperature at 850 hPa (T850), we provide all input fields at 850 hPa. Similarly, when predicting geopotential at 500 hPa (Z500) all input fields are at 500 hPa. 
The years 1999-2013 are used for training, 2014-15 for validation, and 2016-17 for testing. Since the datasets are re-forecasts and a reanalysis, there is no difference in data assimilation and predictions between older and more recent dates, and therefore the selection of consecutive years should not have a major impact. We have verified that the selection of different training, validation and testing splits only has a minor impact on results (e.g., we see 7.6\% improvement with our bias correction network, versus an average of 8.3\% when performing cross-validation). Furthermore, we select these (most recent) years as it is our goal to model and evaluate our networks' capabilities to predict future weather given training-input from the past.
The effects of climate change on the uncertainty of forecasts are currently being explored~\cite{doi:10.1029/2018GL081856}. For our selected parameters and years, these effects are low. 
It is, however, important to use complete years, as different seasons demonstrate different weather patterns. 
We target forecasts with a lead time of 48 hours, and use the reduced ensemble forecasts for 0, 24, and 48 hour lead times as inputs.



\subsection{Data Preprocessing}
\begin{figure}[t]
\centering
\includegraphics[width=.95\linewidth,clip,trim={0.3cm 2cm 0.3cm 0.4cm}]{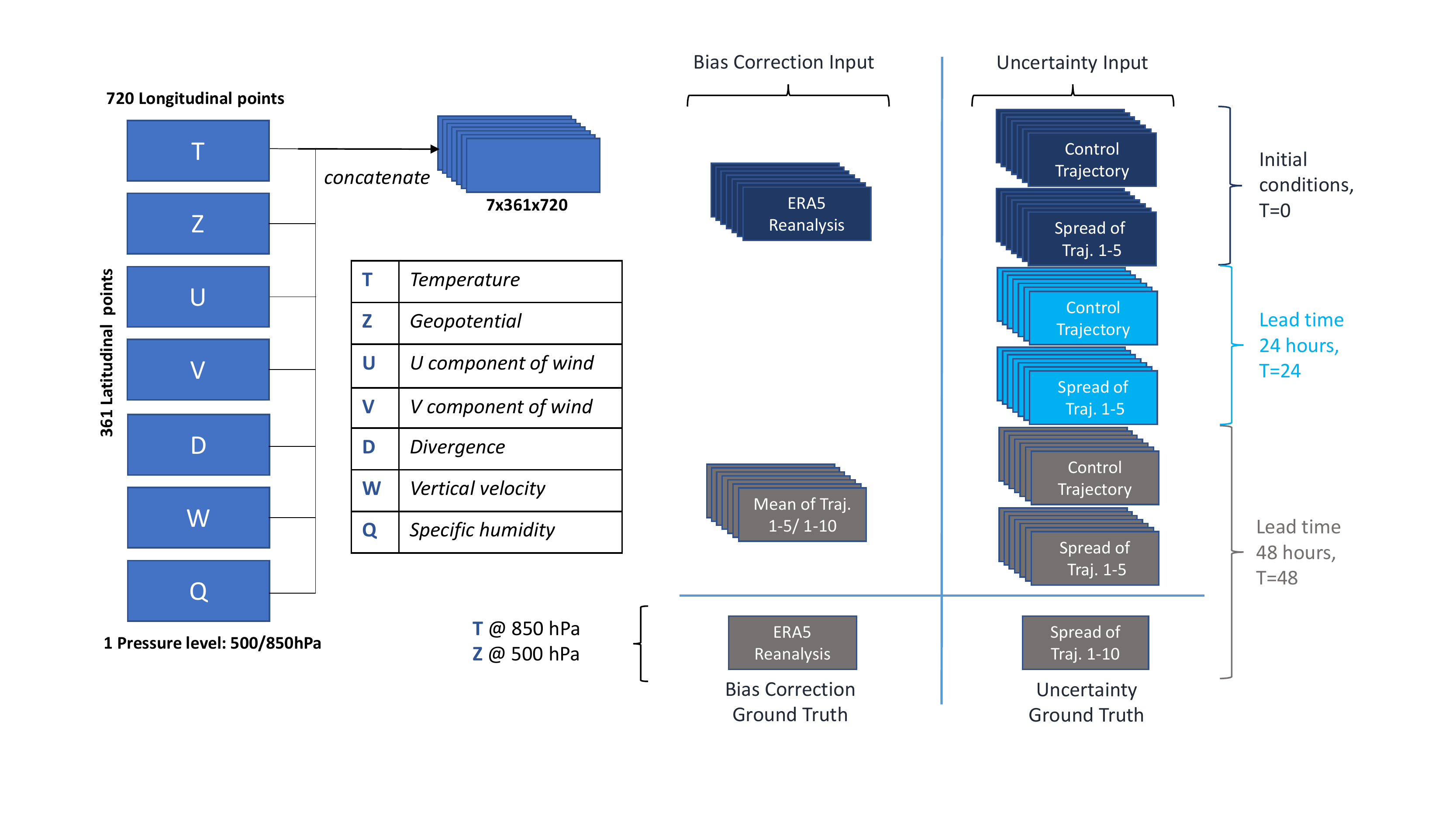}
\caption{Inputs and ground truths to our neural networks.}\vspace{-1em}
\label{fig:Param}
\end{figure}

As the datasets consist of several terabytes of data, we set up a data preprocessing pipeline to enable faster training. 
We first select the relevant inputs and labels to each of our respective models from the data provided in GRIB\cite{grib} format. 
We then convert the data from a 16-bit fixed point format to 32-bit floating point. This simplifies and speeds up training, while not impacting the results.
Finally, we standardize (to zero mean, unit variance) our features and save them in the TFRecord format, which is the preferred dataset file extension in the TensorFlow deep learning framework. The resulting inputs and training targets can be seen in Figure \ref{fig:Param}. 
We base our model inputs on the first five ENS10 trajectories as we observed no significant differences in the average means or spreads when using different selections. 

For DNNs to learn local weather patterns, it is important to keep local spatial differences in variability (coherence) when standardizing meteorological data~\cite{doi:10.1029/2019MS001705}.
However, if only one value per mean and standard deviation are applied to scale values on the whole globe, there will be massive differences for specific regions, e.g., different means and standard deviations closer to the poles compared to the equatorial region. This can lead to poor accuracy when applying CNNs that are translation-invariant. 
At the same time, just applying gridpoint-wise standardization will result in losing important information, otherwise represented through the coherence.

\begin{figure*}[h!]
\centering
\includegraphics[width=0.9\linewidth]{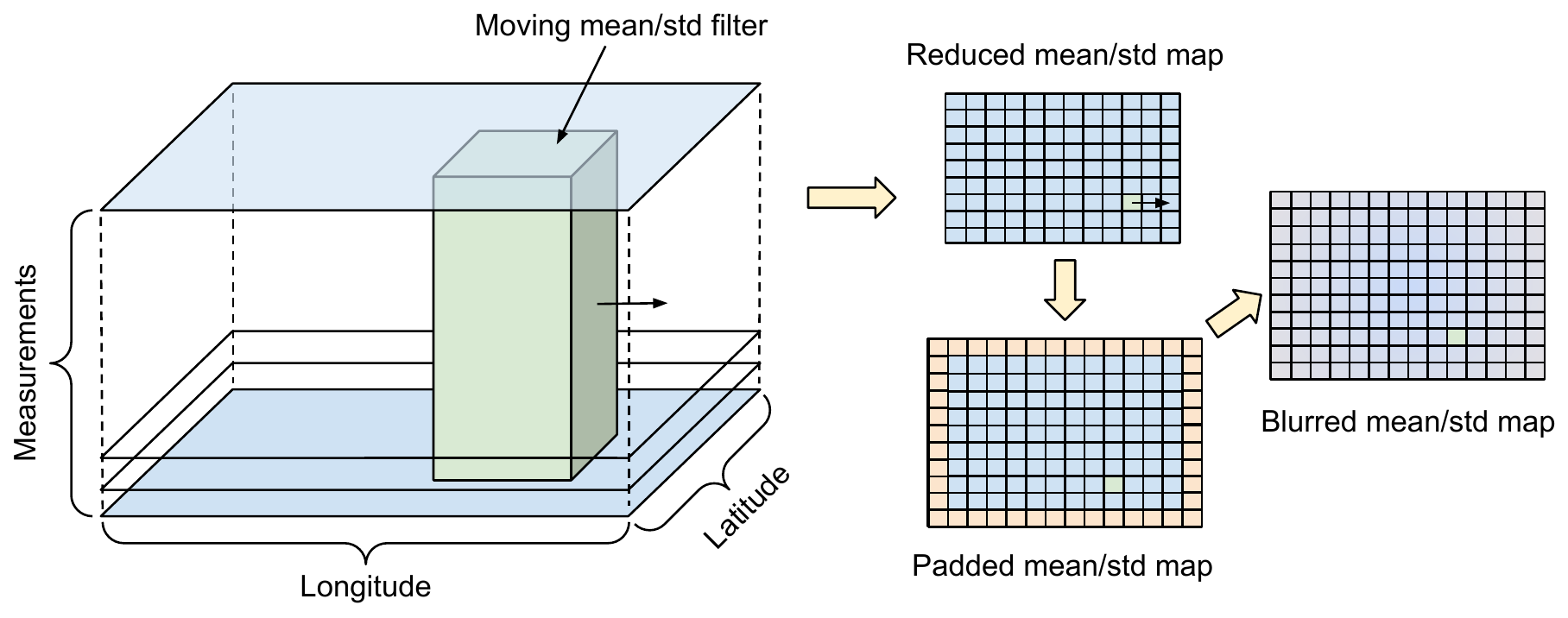}
\caption{Local Area Standardization. The process is done twice, once by taking the mean of the moving filter, and once by taking its standard deviation, thereby obtaining a mean and standard deviation (std) map respectively.}
\label{fig:LAS}
\vspace{-1em}
\end{figure*}

\begin{figure*}[h!]
\centering
\begin{subfigure}{0.49\textwidth}
    \centering
    \includegraphics[width=0.9\linewidth]{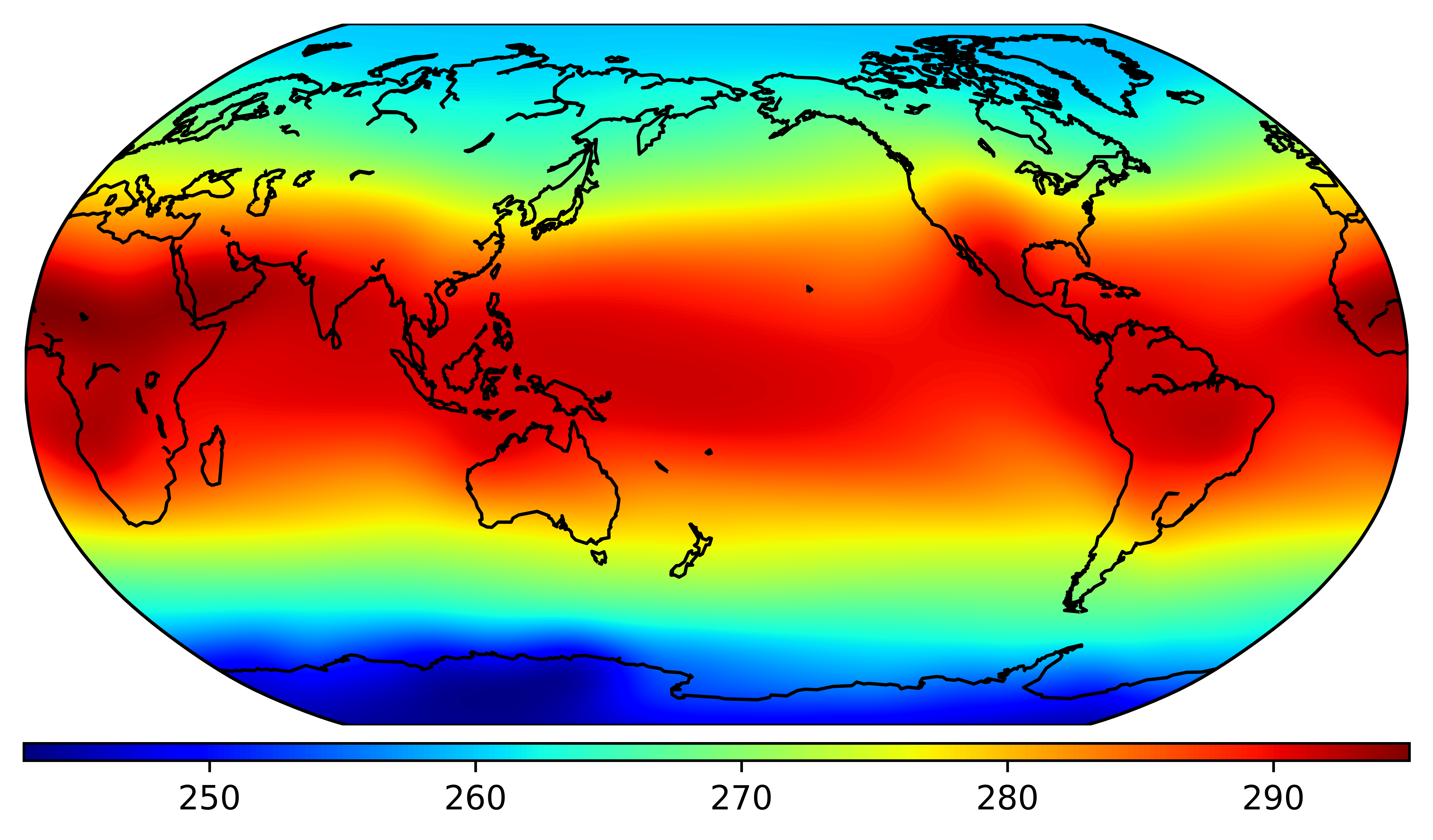}
    \caption{Mean values.}
\end{subfigure}
\begin{subfigure}{0.49\textwidth}
    \centering
    \includegraphics[width=0.9\linewidth]{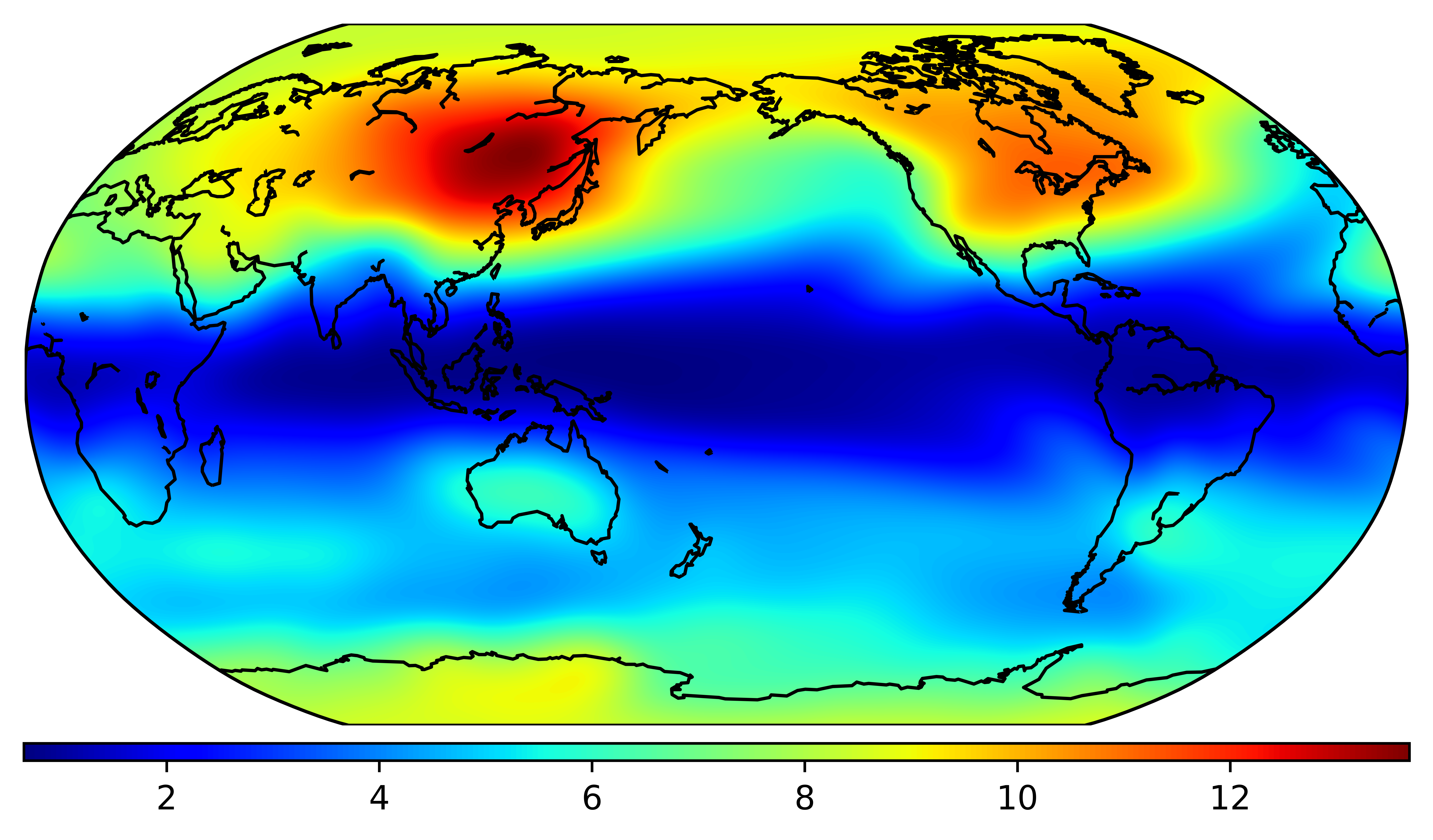}
    \caption{Standard deviation values.}
\end{subfigure}
\caption{LAS values for Temperature at 850 hPa (T850), years 1999-2013 of ENS10.}
\vspace{-1em}
\label{fig:LAS_E}
\end{figure*}

To remedy this problem, we apply a heuristic we refer to as Local Area-wise Standardization (LAS) (see Figures~\ref{fig:LAS} and \ref{fig:LAS_E}).
First we apply a moving average and moving standard deviation filter on our training set. We use a step size of one and a filter size of $7\times7$ (the largest CNN filter size we apply in our DNNs). 
Then, as our mean and standard deviation maps now have reduced dimensions, we pad them using the edge values for latitudes and a wrap around for longitudes. Finally, we apply a Gaussian filter (truncated at four standard deviations) with a large standard deviation (of 10) to the padded result. This upscaling method, first padding and then blurring the upscaled feature map with a gaussian filter, allows for the coherence to be kept between singular grid points. 

Using LAS we notice a relative improvement in our DNN results of around 15\% for spread prediction on our validation sets, as well as faster convergence times compared to applying the same standardization for all grid points. However, as expected, we see no difference when using it with our LCNs, which are not translation-invariant (see Section \ref{sec:nn}\ref{subsec:output-bias-correction}). The method is not fine-tuned and serves as an initial effort to reduce the workload on the neural networks by already accounting for reoccurring patterns, such as higher mean temperatures around the equator and reduced standard deviations. 

\section{Neural Networks}
\label{sec:nn}

We develop separate neural networks for our uncertainty quantification and output bias correction tasks, which are described throughout this section: a residual neural network composed of Inception-style modules~\cite{incep} and a U-Net~\cite{UNET} architecture with an additional locally-connected layer, respectively. While prior work in weather uncertainty prediction~\cite{Wunc} used a 3D U-Net~\cite{cciccek20163d} architecture, the DNNs we develop perform better for our tasks.

Residual connections~\cite{he2016deep} (also called ``skip'' connections) pass unmodified features between layers that are not directly connected to each other, allowing them to be directly used by later layers. Such connections were found to be crucial for our results. Indeed, Chen et al.~\cite{chen2018neural} demonstrated that applying successive residual connections has many similarities to ordinary differential equations. We also considered recurrent neural networks, but do not apply them here due to the short input sequences and lack of improvement in prior work~\cite{Wunc}.


\subsection{Uncertainty Quantification Model}

\begin{figure}[h]
    \centering
    \includegraphics[trim=0 390 0 20,clip,width=1\linewidth]{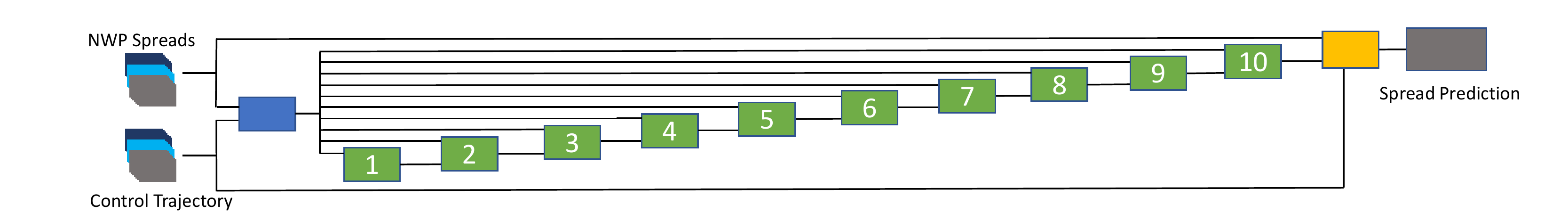}
    \includegraphics[trim=0 0 0 0,clip,width=1\linewidth]{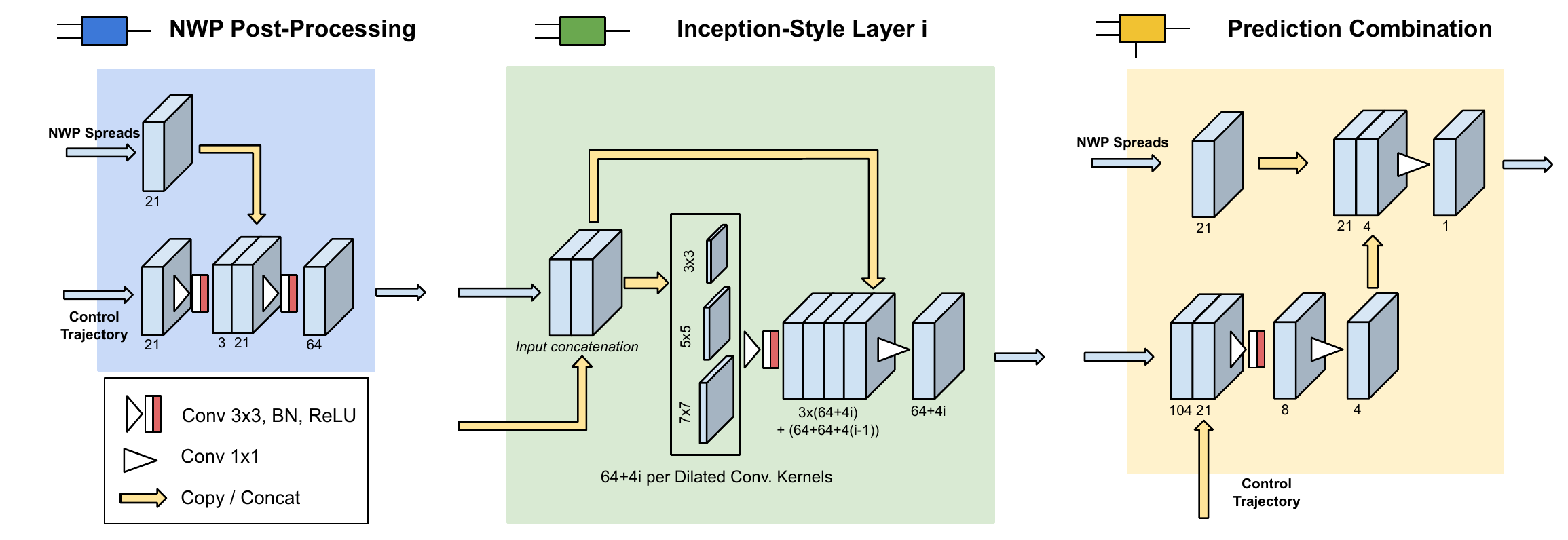}
    \caption{Spread Prediction Network for Uncertainty Quantification. All layers marked \textit{Conv} have a kernel dimension of $1 \times 1$ and are meant to reduce the number of filters. For other convolutional layers we use Batch Normalization (BN) and ReLU activations on the outputs.}
    \label{fig:RN}
\end{figure}

There are many uncertainties present in NWP models and data. Data, or aleatoric, uncertainty stems from observational measurement noise, while model uncertainty comes from structural (e.g., forecast model) and parametric uncertainties. 
In addition to these inherent uncertainties, we also introduce a structural uncertainty by applying a common assumption of NWP that the distribution of errors and uncertainties for meteorological fields, which are represented by the distribution of ensemble members in ensemble predictions, follow Gaussian distributions. 
Our DNN is able to address data and structural uncertainties stemming from NWP; however, we cannot address parametric uncertainties, as the data assimilation pipelines and forecast models we use are fixed and used as prediction labels. 
More specifically, to reduce computational requirements, our DNN initially aims to predict the full ensemble spread using only a subset of NWP ensemble trajectories. The architecture is summarised in Figure~\ref{fig:RN}.

The non-linear nature of our DNN model, which introduces a previously unused structure in NWP and statistical post-processing, combined with the forecasts of the reduced NWP ensemble, helps address the original structural uncertainty. Such a design also allows our model to take into account deterministic forecasts, which it would otherwise struggle to learn.
Additionally, as CNNs are relatively robust to noise, they are naturally able to account for data uncertainty. We also perform a minimal post-processing on NWP output, which reduces the number of parameters by encoding all input features aside from spreads into reduced dimensions. This output is then used for all subsequent steps. 

The core of the DNN is based on the ResNet architecture~\cite{he2016deep}. We use ten \textit{Inception-style modules}~\cite{incep} with residual connections; we did not see any improvement with more layers. Each Inception-style module is composed of three parallel dilated convolutions (where dilation refers to an increased stride between the convolution kernel elements), allowing the network to learn differently-sized receptive fields (local regions). We also perform a channel-wise concatenation of the post-processed NWP output to the input of each Inception-style layer (Figure~\ref{fig:RN}, top). This allows the network to prioritise between different lead times from NWP forecast spreads and its own outputs. Using auxiliary losses for different lead times and depths performed worse than pure NWP predictions.

Finally, the output of the last Inception-style module is combined with the NWP spread through a weighted mean. This guarantees the network performs at least as well as the NWP spread used as input during training. 
By combining this model with our bias correction model and training with ERA5 data as the ground truth (see Section~\ref{sec:nn}\ref{subsec:metrics}), it is possible to also account for parametric uncertainty. This allows our combined networks to cover all types of uncertainty.

\subsection{Output Bias Correction Model}
\label{subsec:output-bias-correction}

\begin{figure}[t]
    \centering
    \includegraphics[trim=0 0.5cm 0 0,clip,width=.95\linewidth]{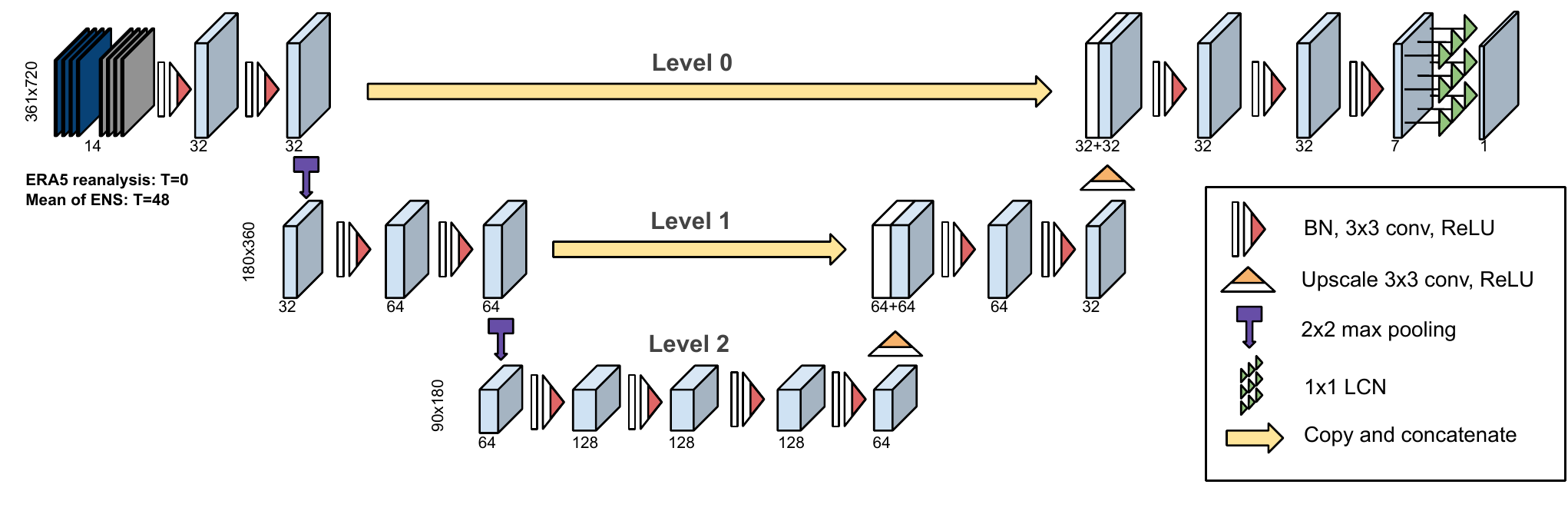}
    \caption{Output Bias Correction model, based on a three-level U-Net and added LCN structure.}
    \vspace{-1em}
    \label{fig:Unet}
\end{figure}

Our output bias correction model, summarised in Figure~\ref{fig:Unet}, corrects for weather-dependent, local biases in NWP forecasts. It is trained using the mean ensemble predictions with a 48-hour lead time and ERA5 data as ground-truth. Since the forecast can resemble the ground-truth, a straightforward predictor will closely resemble the identity function. Prior research~\cite{he2016deep} suggests that approximating an identity mapping with several non-linear layers is difficult. We therefore train our model to predict the difference between the NWP prediction and the ground-truth.

The network is based on a U-Net structure, which repeatedly convolves and downscales inputs, followed by similarly upscaling the features (i.e., forming a ``U'' shape, as seen in Figure~\ref{fig:Unet}). Specifically, each scaling operation is applied after several layers of convolution. Residual connections are also used between the down- and upscaling sides. We make three key changes to adapt the standard U-Net to our task. First, instead of up-convolution, we use bilinear interpolation to upscale, followed by a $3 \times 3$ convolution with stride 1. This is due to checkerboard artifacts that are known to appear when only using a simple deconvolution operation~\cite{odena2016deconvolution}. Second, we reduce the number of levels in the U-Net, from five levels to three, as we found using additional levels resulted in overfitting on our data. Finally, we reduced the number of filters in each convolution by half, as we observed no additional performance improvements by using more.

As we aim to predict the bias emerging from specific regional patterns, the translational invariance of regular convolution hinders performance. We therefore use a locally-connected network (LCN) as the last layer. LCNs perform a similar operation to regular convolution, but instead of sharing filters across all spatial points, independent filters are used for each output. When training, we apply $\ell_1$ regularization on the difference between all adjacent filters in an LCN, to encourage adjacent filters to learn similar weights; for an infinite regularization parameter, the LCN converges to a convolutional layer. This helps avoid overfitting. 

In order to remain computationally efficient, our best models first use a U-Net to perform feature extraction and then apply an LCN to obtain our final output bias correction. As the U-Net is able to learn long-range dependencies, a single LCN with $1 \times 1$ kernels is sufficient to learn the gridpoint-wise dependencies, and we observed no improvements by using larger filter sizes.

\subsection{Metrics}
\label{subsec:metrics}

As our models solve different tasks, we need to use different metrics to evaluate them. When training, we treat both uncertainty quantification and output bias correction as regression problems, and aim to predict extreme cases (outliers). The ENS10 spread is used as ground truth for the uncertainty network, while the ERA5 values are used for the bias correction. Initially, we train both networks on the mean-squared error (MSE) and evaluate them with root mean-squared error (RMSE).
However, when predicting the spread, the results lack the sharp edges that exist in the original forecasts. 
In computer vision, this problem is mitigated using the Structural SIMilarity (SSIM) metric~\cite{1284395}. There can be infinitely many solutions to the task of minimising RMSE. While remaining within the realm of these solutions, the SSIM measures the structural similarity between two images, with $1.0$ being a perfect match. 
Therefore, we switch to using the negative mean SSIM of our prediction compared to the full ENS10 ensemble as our training loss for the uncertainty quantification model.

\begin{wrapfigure}{r}{0.5\textwidth} 
\vspace{-20pt}
\centering
\includegraphics[width=1\linewidth]{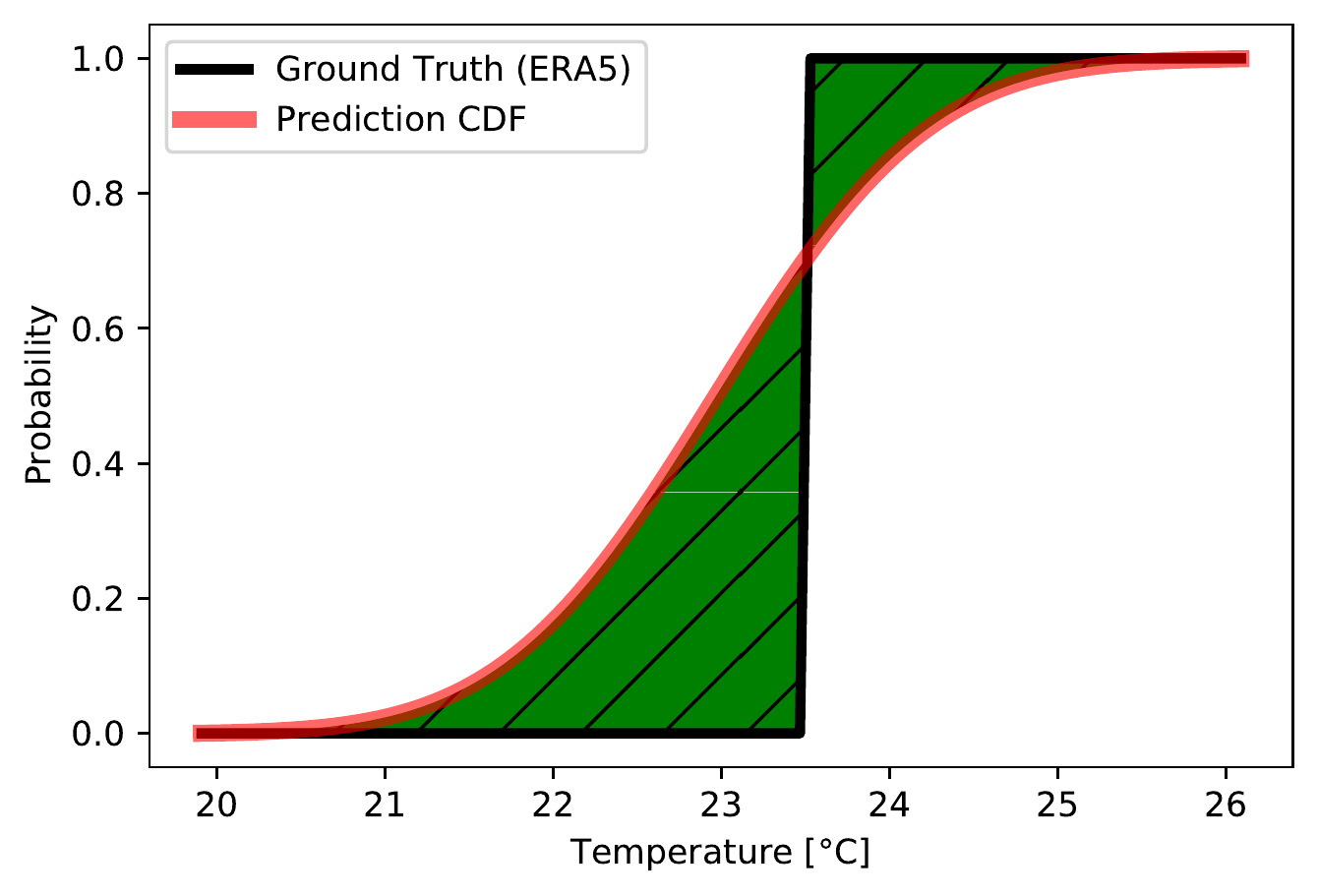}
\caption{Visualisation of CRPS for a temperature prediction. CRPS is calculated as the integral of the square of the green area.}
\label{fig:CRPS_AREA}
\vspace{-10pt}
\end{wrapfigure}

To then gain an understanding of the forecast skill of our combined predictions and of the ENS10 forecasts, we use the Continuous Ranked Probability Score (CRPS)\cite{doi:10.1175/1520-0434(2000)}. 
CRPS, generally used to measure whether ensemble methods represent uncertainty correctly, is the integral of the square of the difference between the Cumulative Distribution Function (CDF) of the probabilistic predictions $F$ and the ground truth $y$ (see Figure \ref{fig:CRPS_AREA}):
$$
\text{CRPS}(F,y)=\int_{-\infty}^{\infty}[F(x)-\textbf{1}_{x>y}]^{2}dx
$$

Here, $\textbf{1}_{x>y}$ is the indicator function (equivalent to the CDF of a deterministic value).
In the following, we do not only combine the uncertainty and bias correction networks to calculate CRPS, but also perform PDF calibration by training a combination of both in a network that is optimised to minimise the CRPS. 
We achieve this by replacing the labels of our uncertainty quantification network, which were previously the spread values of the full ENS10 trajectories, by the difference between the ground truth and the output bias corrected forecast $\Delta{P}$. 
With our assumption of the forecasts being of a Gaussian distribution, $\Delta{P}$, the error function $\Phi$ and standard deviation $\sigma$ we then set the CRPS loss as follows (see Appendix~\ref{sec:CRPS_Equation} for full derivation):
\begin{align*}
\Delta{P} &= \text{Ground Truth} - \text{Prediction}\\
\Phi(x) &= \frac{2}{\sqrt{\pi}}\int_{0}^{x}e^{-t^2}dt\\
\text{CRPS}(\sigma,\Delta{P})&=\Delta{P}\Phi\left(\frac{\Delta{P}}{\sqrt{2}\sigma}\right) + \frac{\sigma}{\sqrt{\pi}}(-1+\sqrt{2}e^{-\frac{\Delta{P^2}}{2\sigma{^2}}}).
\end{align*}

Finally, the relative CRPS improvement of a prediction over the original raw ensemble is defined as the Continuous Ranked Probability Skill Score (CRPSS):
$$
\text{CRPSS}(\text{CRPS}_{pred},\text{CRPS}_{orig})=1-\frac{\text{CRPS}_{pred}}{\text{CRPS}_{orig}}.
$$

\subsection{Implementation}

Our networks are implemented with the TensorFlow deep learning framework~\cite{tensorflow2015-whitepaper}. Layers are initialised with a truncated normal distribution. We train with the Adam optimiser~\cite{kingma2014adam} with a learning rate of 0.001 and $\ell_2$ regularization. Our models have not been fine-tuned extensively, and there is further potential for improvement. More implementation details can be found in our GitHub repository.

The uncertainty networks are trained for 4,725 update steps with a batch size of 2, requiring about four hours on one Nvidia V100 32 GB GPU. The bias correction networks are trained for the same wall-clock time, taking about 25,000 update steps with a batch size of 2. We use early stopping, i.e., ending the training process once the validation loss stops decreasing, to identify the best parameters. Training can be done once and the resulting networks used until the ensemble prediction system is upgraded. Using the same GPU, inference for one parameter and forecast on a global grid takes approximately 0.31 seconds per network.

\section{Results}
\label{sec:results}

\begin{table}[h]
\centering
\begin{tabular}{ll}
\specialrule{.1em}{.05em}{.05em} 
\textbf{Notation} & \textbf{Description}         \\ \hline
B\{$n$\}          & Output bias correction NN trained with $n$ trajectories        \\
U\{$n$\}          & Spread prediction NN trained with $n$ trajectories        \\
E\{$n$\}          & Ensemble with $n$ trajectories \\
Lin\{$n$\}        & Gridpoint-wise linear regression from $n$ trajectories \\
C               & Uncertainty NN trained on CRPS \\ 
G               & Ground truth data from ERA5 \\ \specialrule{.1em}{.05em}{.05em}
\end{tabular}
\caption{Notation for our model configurations and ground-truth data.}
\label{tab:Mdesc}
\vspace{-1em}
\end{table}

We primarily train our models to predict T850 but also evaluate their prediction capacity on Z500. 
First, our uncertainty quantification and bias correction networks are evaluated separately on the global RMSE for the spread of ENS10 forecasts or the ERA5 ground-truth respectively. All results are for a forecast lead time of 48 hours. In addition to our DNNs, we train linear regression models on ensemble trajectories as another baseline (see Table~\ref{tab:Mdesc}).

\begin{figure*}[h]
\centering
\begin{subfigure}{0.24\textwidth}
    \centering
    \includegraphics[height=1.6in]{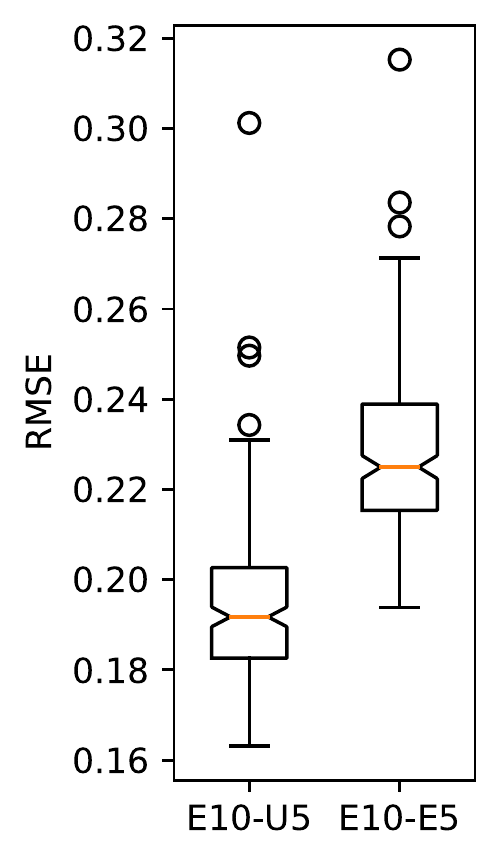}
    \caption{T850}
\end{subfigure}
\begin{subfigure}{0.24\textwidth}
    \centering
    \includegraphics[height=1.6in]{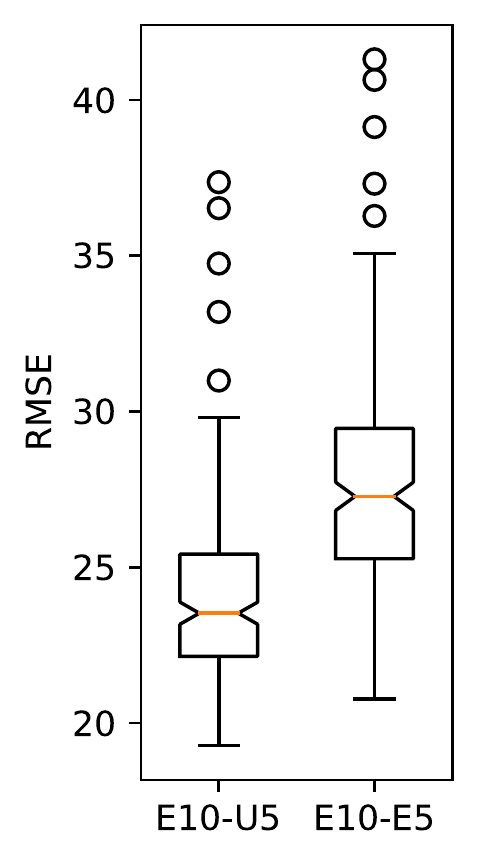}
    \caption{Z500}
\end{subfigure}
\begin{subfigure}{0.24\textwidth}
    \centering
    \includegraphics[height=1.6in]{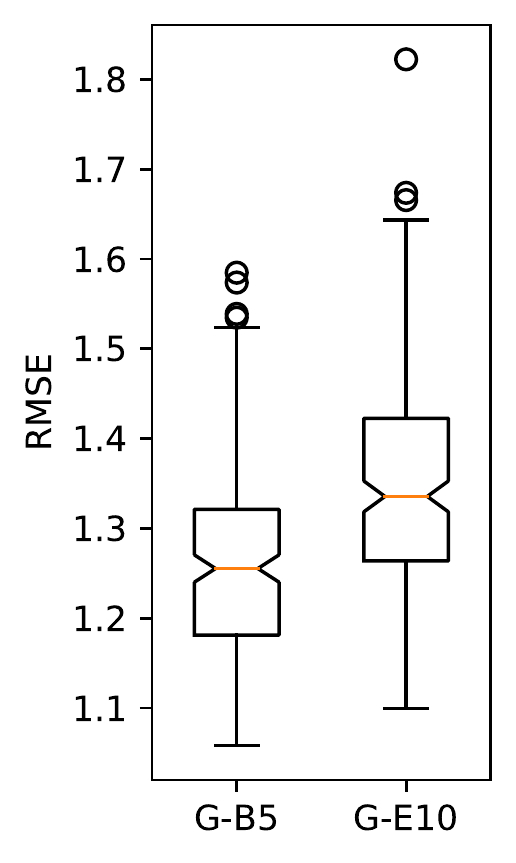}
    \caption{T850}
\end{subfigure}
\begin{subfigure}{0.24\textwidth}
    \centering
    \includegraphics[height=1.6in]{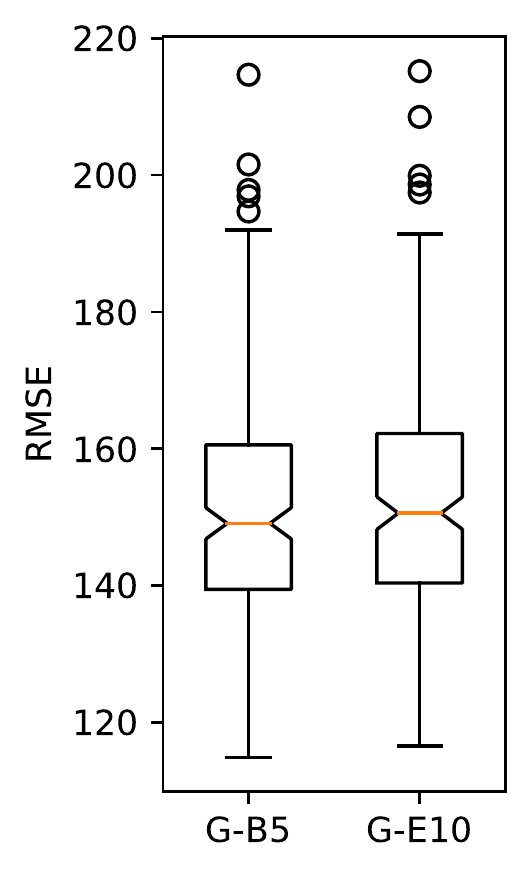}
    \caption{Z500}
\end{subfigure}
\caption{Notched boxplots for the global RMSE of the Uncertainty Quantification and Output Bias Correction Networks each day of our test set (2016-17). For the x-axis $\text{"A}-\text{B"} \mathrel{\widehat{=}} \text{RMSE}(A,B)$.}
\vspace{-1em}
\label{fig:SComp}
\end{figure*}

Figures~\ref{fig:SComp} (a, b) show the improvement in spread prediction of our uncertainty network using five ensemble trajectories, compared to simply using the five trajectories. There are significant improvements for both temperature and geopotential. Figures~\ref{fig:SComp} (c, d) show our output bias correction for predicting deviating weather patterns given a forecast mean and no measure of uncertainty. We see improvements for T850, but our network does not provide a strong global improvement for Z500, which we analyze more thoroughly through case studies (see Section~\ref{sec:results}\ref{subsec:extreme-weather}).


\begin{table}[h]
  \centering
  \begin{tabular}{c c c c c c c c c c c c}
    \specialrule{.1em}{.05em}{.05em} 
    T850 & E3 & E4 & E5 & E6 & E7 & E8 & E9 & Lin5 & U3 & U4 & U5 \\ \hline
    Abs. & 0.35 & 0.28 & 0.23 & 0.19 & 0.15 & 0.11 & 0.07 & 0.21 & 0.26 & 0.23 & 0.19 \\
    Rel. & - & - & - & - & - & - & - & 8.9\% & 26.6\% & 18.7\% & 16.4\% \\
    \specialrule{.1em}{.05em}{.05em} 
  \end{tabular}
  \caption{T850 average RMSE towards E10 spread for different ensemble sizes and models for the test set (2016-17). Abs.: Absolute rounded values. Rel.: Relative improvement over the original forecast.}
  \label{table:UNC_T}
\end{table}

\begin{table}[h]
  \centering
  \scriptsize
  \begin{tabular}{c c c c c c c c c}
    \specialrule{.1em}{.05em}{.05em} 
    Parameter & Lin10 & UN0 & UN1 & UN2 & UN0-LCN & UN1-LCN & UN2-LCN & UN1-LCN-reg \\ \hline
    T850 & 4.8\% & 6.3\% & 7.1\% & 6.7\% & 7.6\% &  7.7\% & 7.6\% & \textbf{7.9\%} \\
    Z500 & 2.1\% & 1.2\% & 1.6\% & 1.0\% & \textbf{2.6}\% & 2.5\% & 2.4\% & 2.3\% \\
    \specialrule{.1em}{.05em}{.05em} 
  \end{tabular}
  \caption{Ablation study of output bias correction, measured by the relative RMSE improvement over the ensemble mean forecast on the test set. UN$i$: U-Net with $i$-levels. UN$i$-LCN: U-Net with $i$-levels followed by an LCN. *-reg: with $\ell_1$ regularization ($\lambda=10$) on the LCN.}
  \vspace{-1em}
  \label{table:bias_correction result}
\end{table}

Table \ref{table:UNC_T} shows our analysis of the final results for the uncertainty quantification network using different numbers of trajectories on T850. 
With a low number of trajectories we can see that our model is already able to predict uncertainty to some degree. 
We observe diminishing returns in the relative performance as the number of trajectories increases, due to the rapidly growing proportion of trajectories out of the full ensemble that is used. However, the networks still consistently outperform the reduced ensemble baselines. This is even the case for a relatively high number of trajectories (half). 
We also present the results of an ablation study --- exploring the impact of different structures and features --- on the output bias correction networks in Table \ref{table:bias_correction result}. 
The results show that T850 correction benefits from convolution operations, whereas Z500 correction benefits most from using a locally connected structure. Further supporting this, we observe that $\ell_1$ regularization does not have as significant an impact for Z500, suggesting that local, independent filters are important for prediction.

To understand the contribution of the input fields to the result, we perform another ablation study, training networks with the predicted field as the only input. The results are listed in Table \ref{table:data_ablation_result}, confirming the importance of using multiple input fields to provide more accurate predictions.

\begin{table}[h]
  \centering
  \begin{tabular}{l l c c}
  \toprule
  Network & Predicted & \multicolumn{2}{c}{Input Fields}\\
  & Parameter & Predicted Only & All Fields\\
 \midrule
 \multirow{2}{*}{Uncertainty Quantification} & T850 & 12.3\% & 16.4\%\\
 & Z500 & 12.8\% & 12.9\%\\
 \multirow{2}{*}{Bias Correction} & T850 & 6.90\% & 7.59\%\\
 &Z500 & 2.06\% & 2.37\%\\
 \bottomrule
 \end{tabular}
  \caption{Ablation study of input fields and the resulting relative improvement for the uncertainty quantification and bias correction (UN1-LCN) networks.}
  \vspace{-1.5em}
  \label{table:data_ablation_result}
\end{table}

\begin{figure}[t]
\centering
\begin{subfigure}{.49\textwidth}
    \centering
    \includegraphics[width=1.\linewidth]{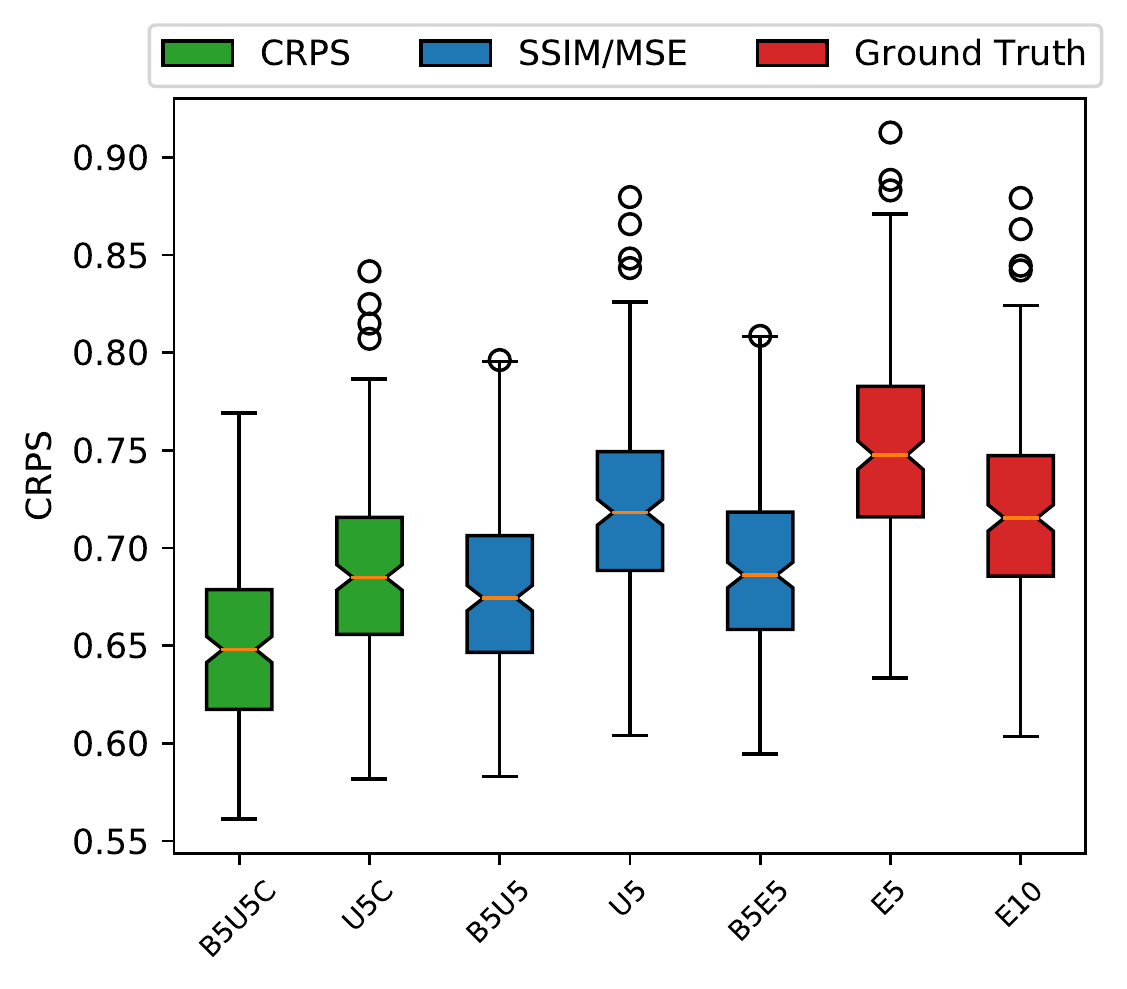}
    \caption{T850}
\end{subfigure}
\begin{subfigure}{.49\textwidth}
    \centering
    \includegraphics[width=1.\linewidth]{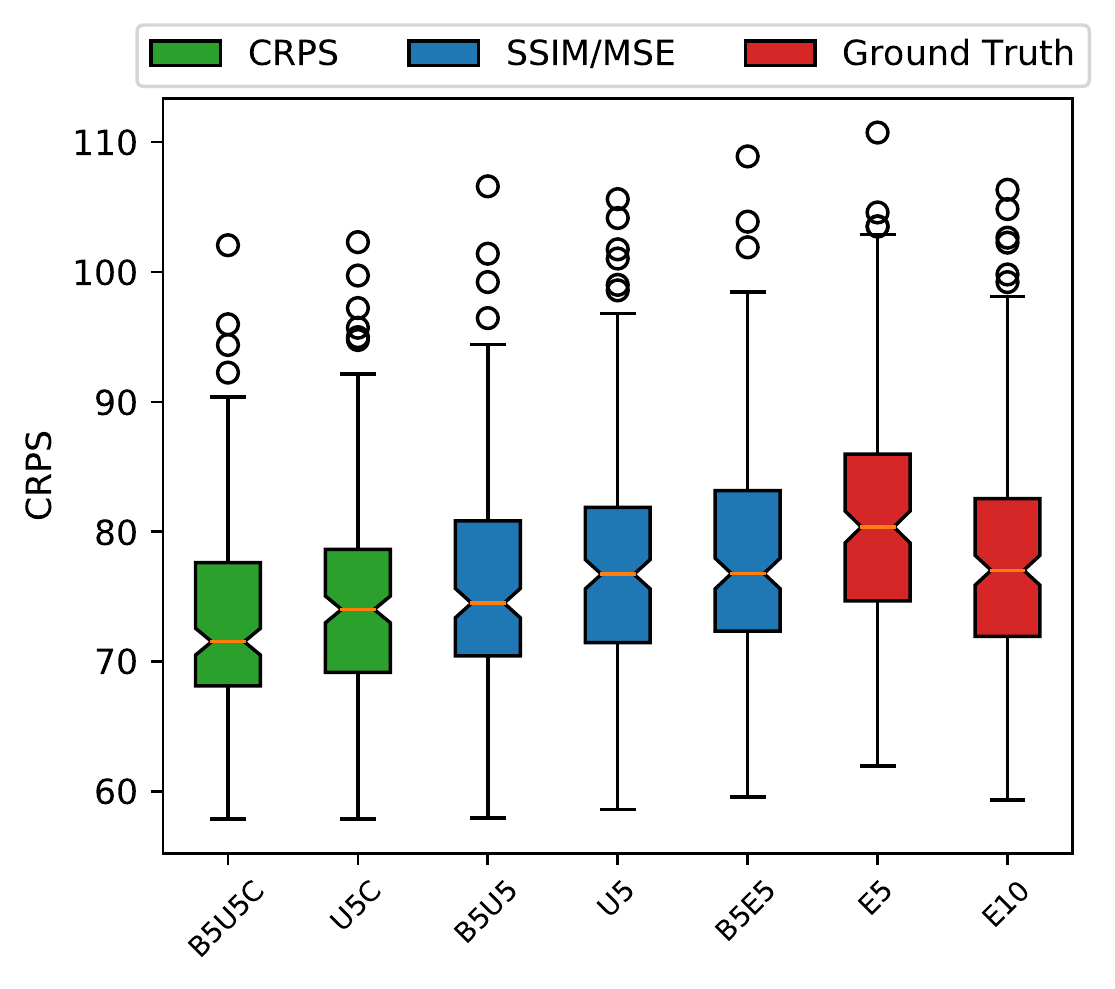}
    \caption{Z500}
\end{subfigure}
\caption{Notched boxplots for the global average CRPS values for each day in our test set (2016-17), for all our networks and raw ensemble combinations.}
\vspace{-1em}
\label{fig:CRPS}
\end{figure}

RMSE is insufficient to measure probabilistic forecast skill, as it does not encompass both mean and spread. We therefore also consider CRPS (lower scores are better). We use E5 and E10, the raw five- and ten-member ensembles, as our baselines.
The results are presented in Figure \ref{fig:CRPS}. 
\begin{wraptable}{r}{6cm}
\vspace{-1em}
\begin{tabular}{lll}
\specialrule{.1em}{.05em}{.05em} 
\textbf{CRPSS} & \textbf{Z500} & \textbf{T850}         \\ \hline
B5U5C towards E10          & 0.0756 & 0.1098      \\
B5U5C towards E5           & 0.1074 & 0.1458      \\
\specialrule{.1em}{.05em}{.05em} 
\end{tabular}
\caption{Continuous Ranked Probability Skill Scores over our test set (2016-17).}
\label{tab:CRPSS}
\vspace{-1em}
\end{wraptable}
To measure our improvements, we specifically focus on the CRPSS of our PDF calibration network trained on CRPS, as compared to the raw ensemble outputs in Table \ref{tab:CRPSS}.
However, both of the models, trained on CRPS and SSIM, outperform the full ensemble forecast CRPS for both T850 and Z500, despite the smaller number of ensemble members that are used. The major source of improvement for our DNNs, especially for T850, is a reduction in extreme values (outliers), which are indicative of forecast busts. We are probably able to achieve better improvements for T850 when compared to Z500 as the temperature field is likely to show more significant biases.

As another baseline comparison, we run EMOS~\cite{doi:10.1175/MWR2904.1} with the first 5 ensemble members of ENS10 as input. There are 8 parameters in this formulation, and they are trained to minimise CRPS using the ERA5 dataset as the ground truth. We use the same training/validation/test set split and training algorithm (Adam with early stopping) as our DNNs. For T850, we observe a resulting CRPS improvement of 5.5\% with EMOS over the test set, compared with 14.5\% for B5U5C.

\subsection{Extreme Weather Forecasts}
\label{subsec:extreme-weather}

\begin{figure*}[t]
\centering
\begin{subfigure}{0.24\textwidth}
    \centering
    \includegraphics[width=1\linewidth]{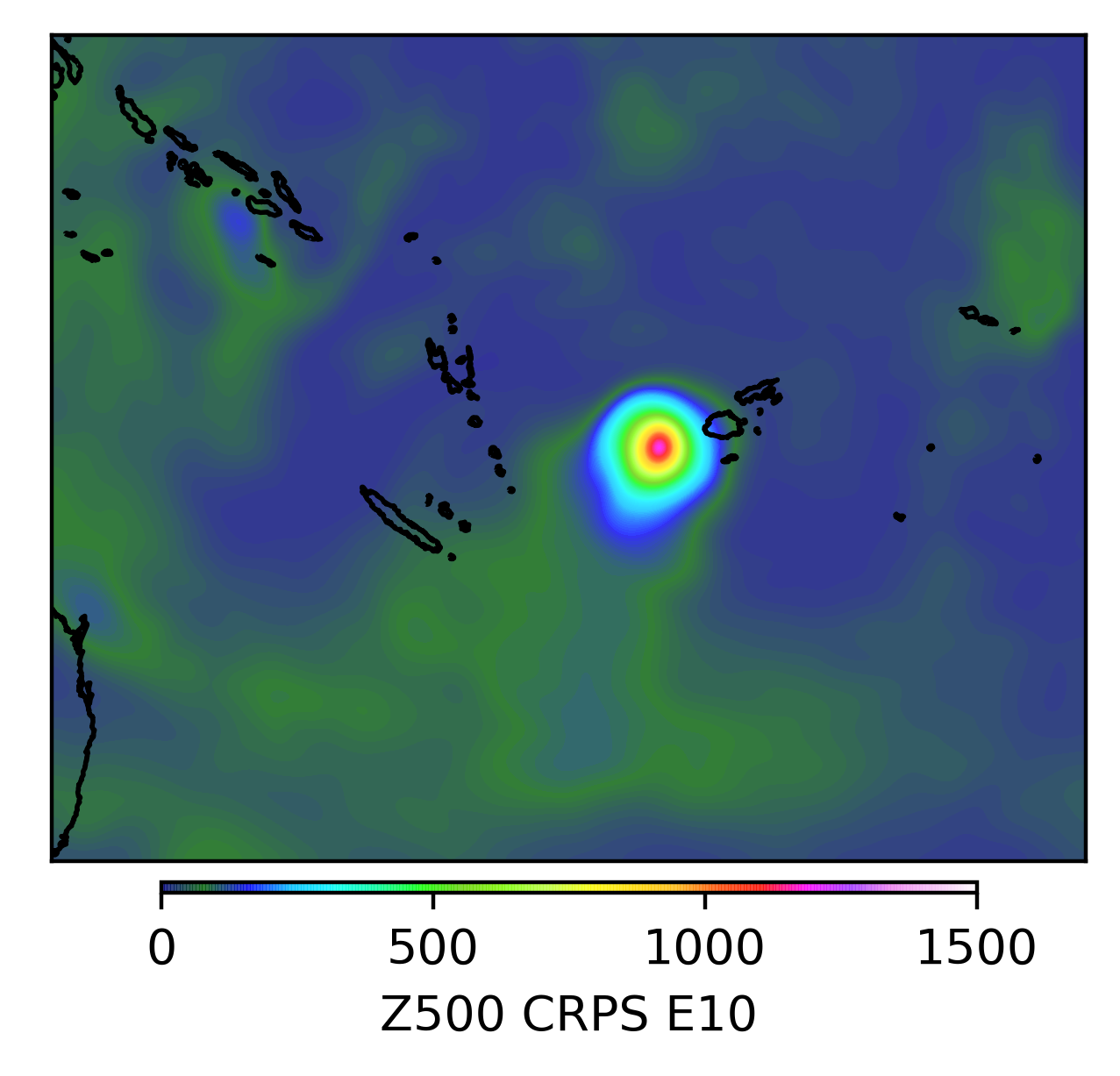}
    \caption{E10}
\end{subfigure}
\begin{subfigure}{0.24\textwidth}
    \centering
    \includegraphics[width=1\linewidth]{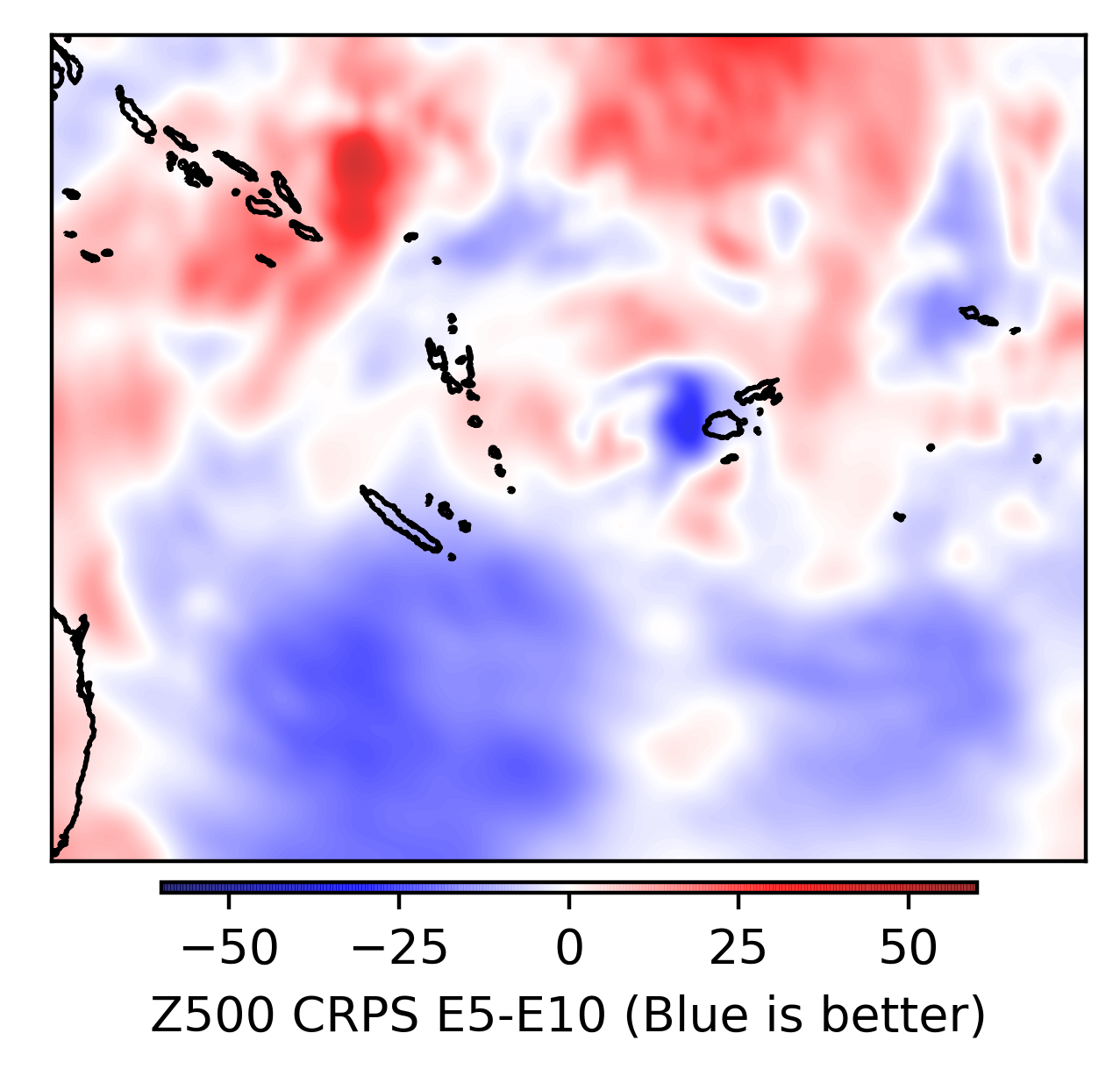}
    \caption{E5-E10}
\end{subfigure}
\begin{subfigure}{0.24\textwidth}
    \centering
    \includegraphics[width=1\linewidth]{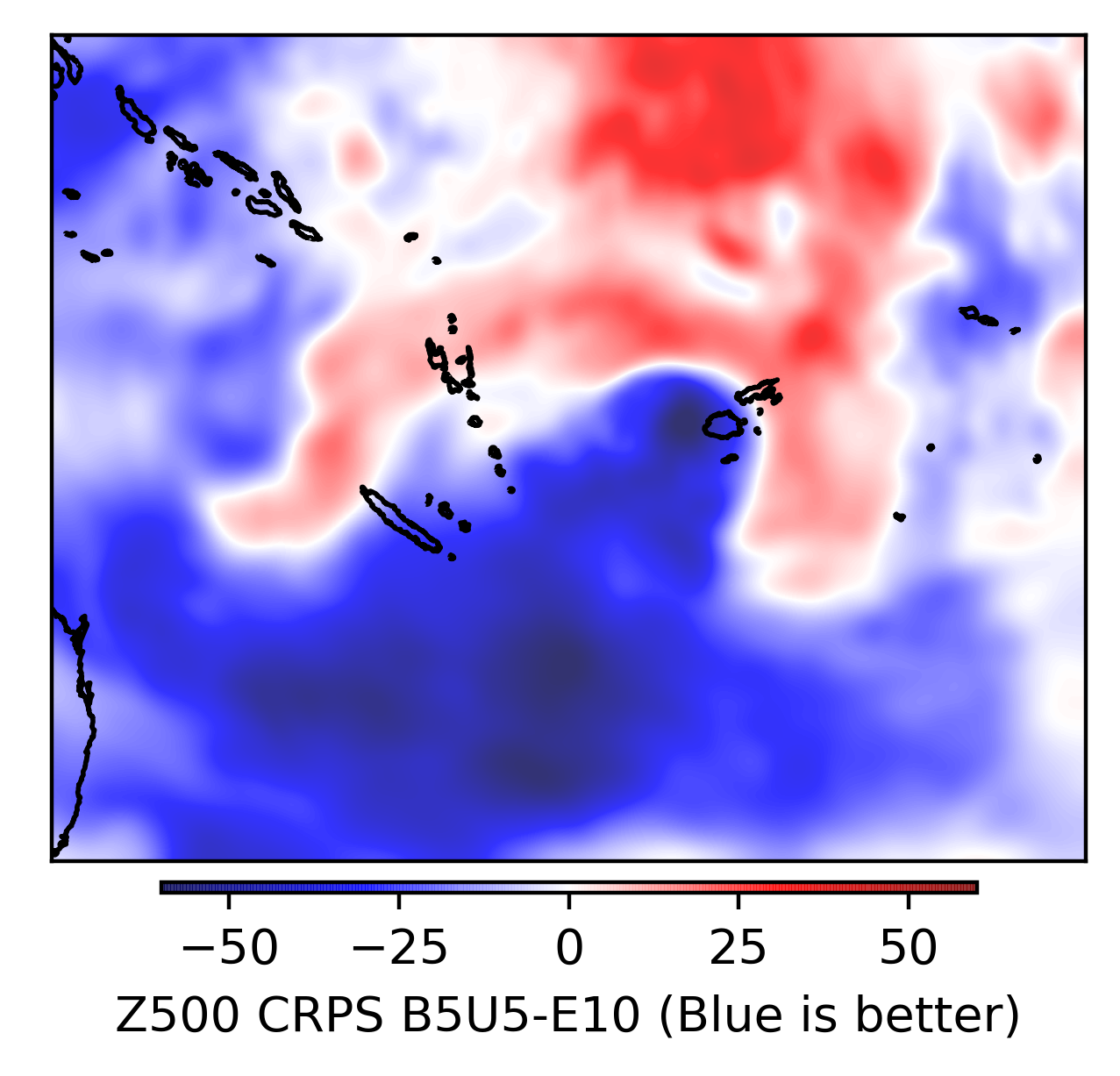}
    \caption{B5U5-E10}
\end{subfigure}
\begin{subfigure}{0.24\textwidth}
    \centering
    \includegraphics[width=1\linewidth]{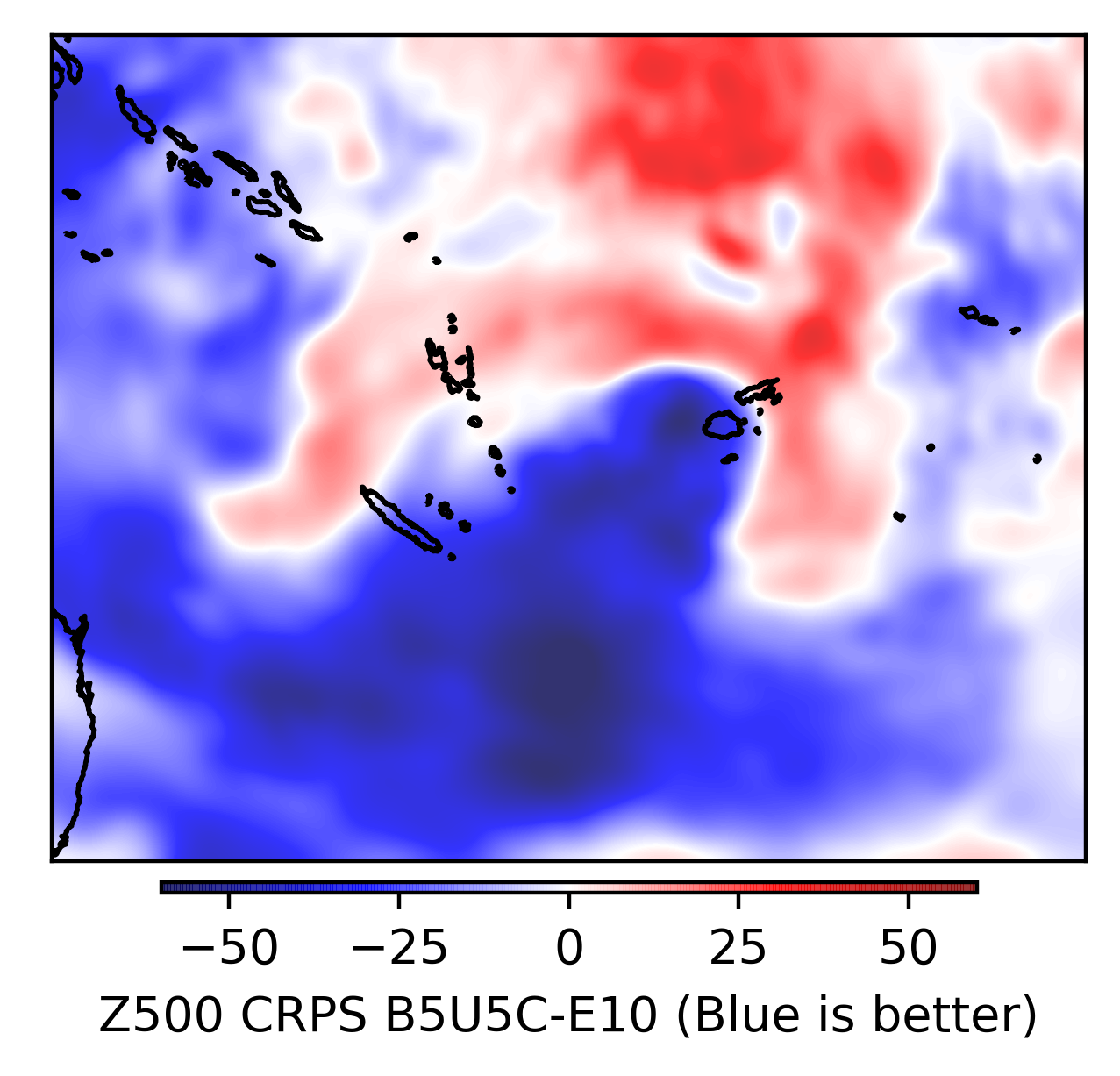}
    \caption{B5U5C-E10}
\end{subfigure}
\caption{Tropical cyclone Winston, raging from February until March 2016 over Fiji, has been classified as the most intense cyclone in the southern hemisphere ever recorded, according to the Southwest Pacific Enhanced Archive for Tropical Cyclones. It reached category five on February 20. We present the prediction for Z500 as forecast on February 19 for February 21, and differences in CRPS. (a) The CRPS for the ten-member ensemble. The centre of the cyclone is clearly visible. (b)-(d) The difference in CRPS between the ten-member ensemble and five-member ensembles with and without post-processing.
Our CRPS-trained network shows improvement over E10. It also demonstrates similar confidence in the southern area where the cyclone is moving, while being worse where the cyclone has already passed. This results in a large forecast skill improvement in the selected area, with a CRPSS of 0.261 (26.1\% improvement over E10).}
\label{fig:Winston}
\end{figure*}

\begin{figure*}[t]
\centering
\begin{subfigure}{0.24\textwidth}
    \centering
    \includegraphics[width=1\linewidth]{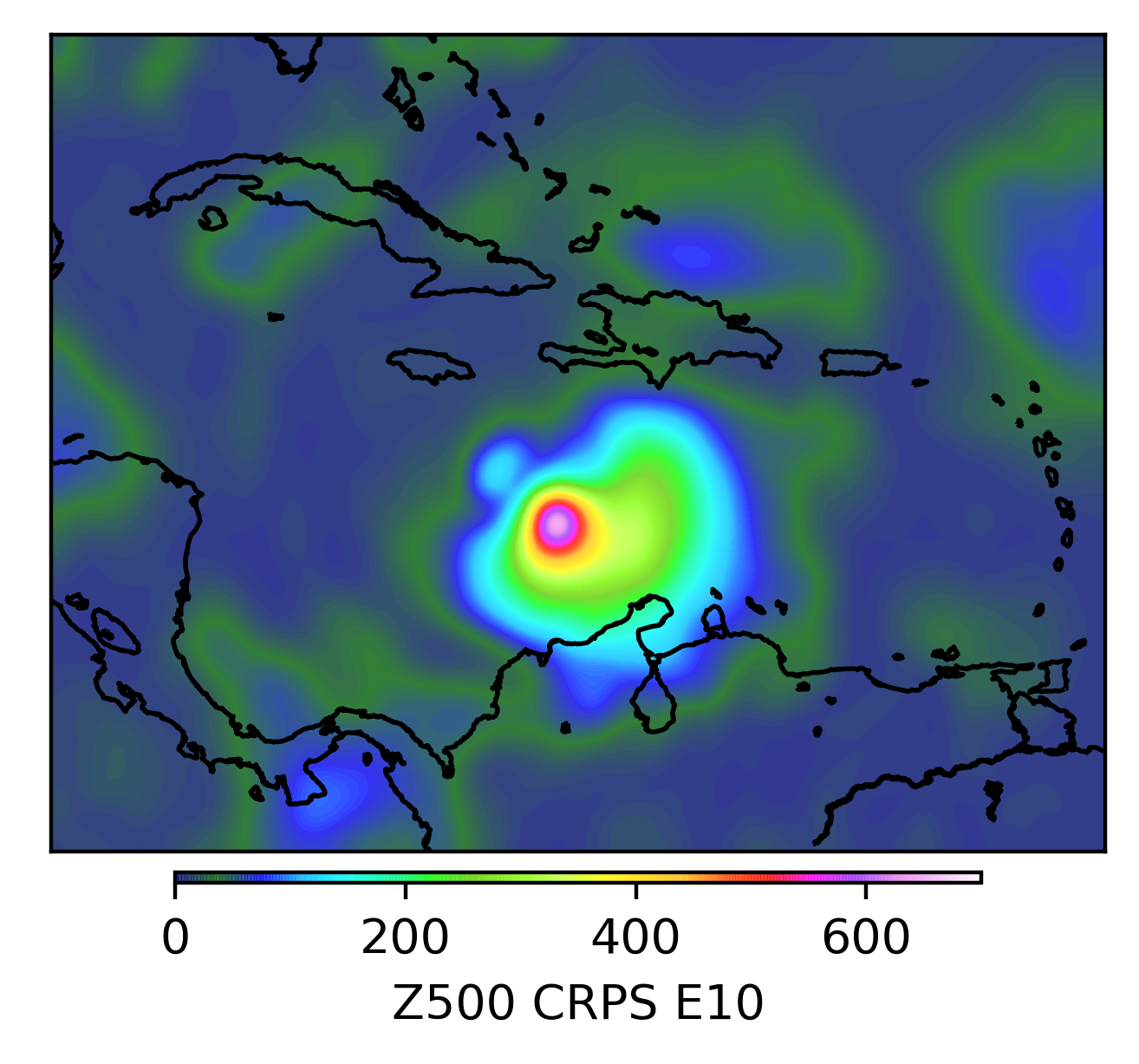}
    \caption{E10}
\end{subfigure}
\begin{subfigure}{0.24\textwidth}
    \centering
    \includegraphics[width=1\linewidth]{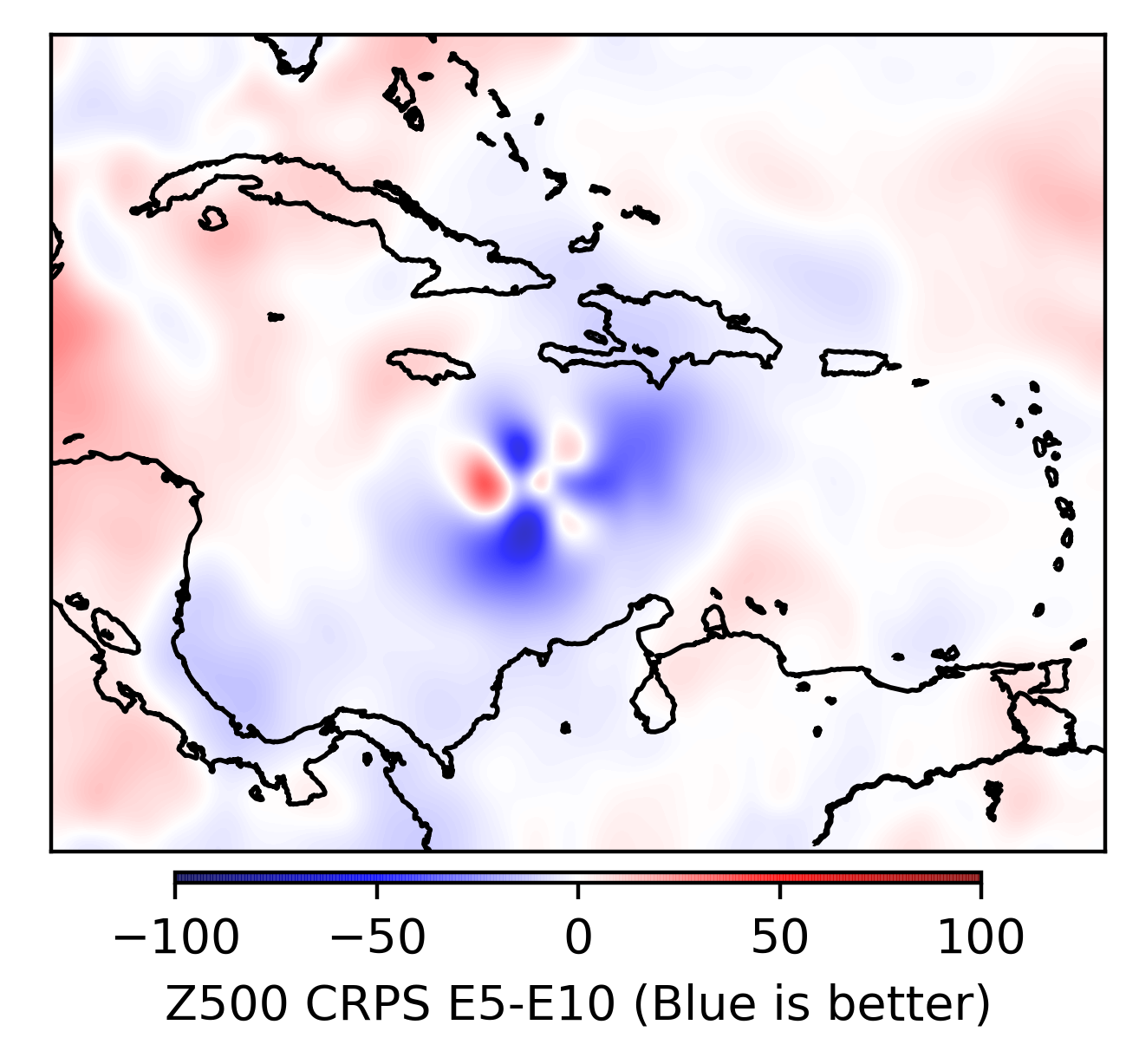}
    \caption{E5-E10}
\end{subfigure}
\begin{subfigure}{0.24\textwidth}
    \centering
    \includegraphics[width=1\linewidth]{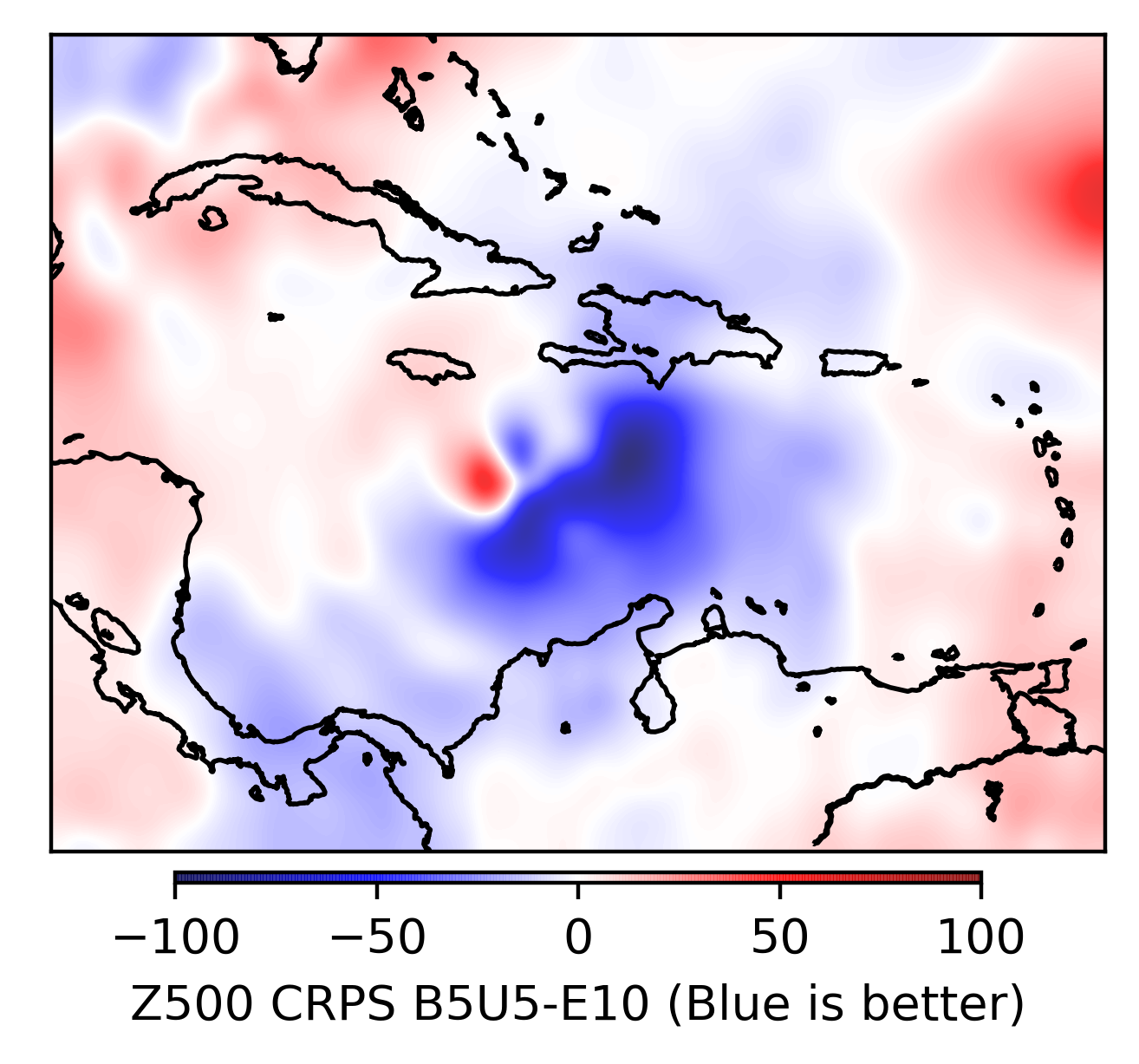}
    \caption{B5U5-E10}
\end{subfigure}
\begin{subfigure}{0.24\textwidth}
    \centering
    \includegraphics[width=1\linewidth]{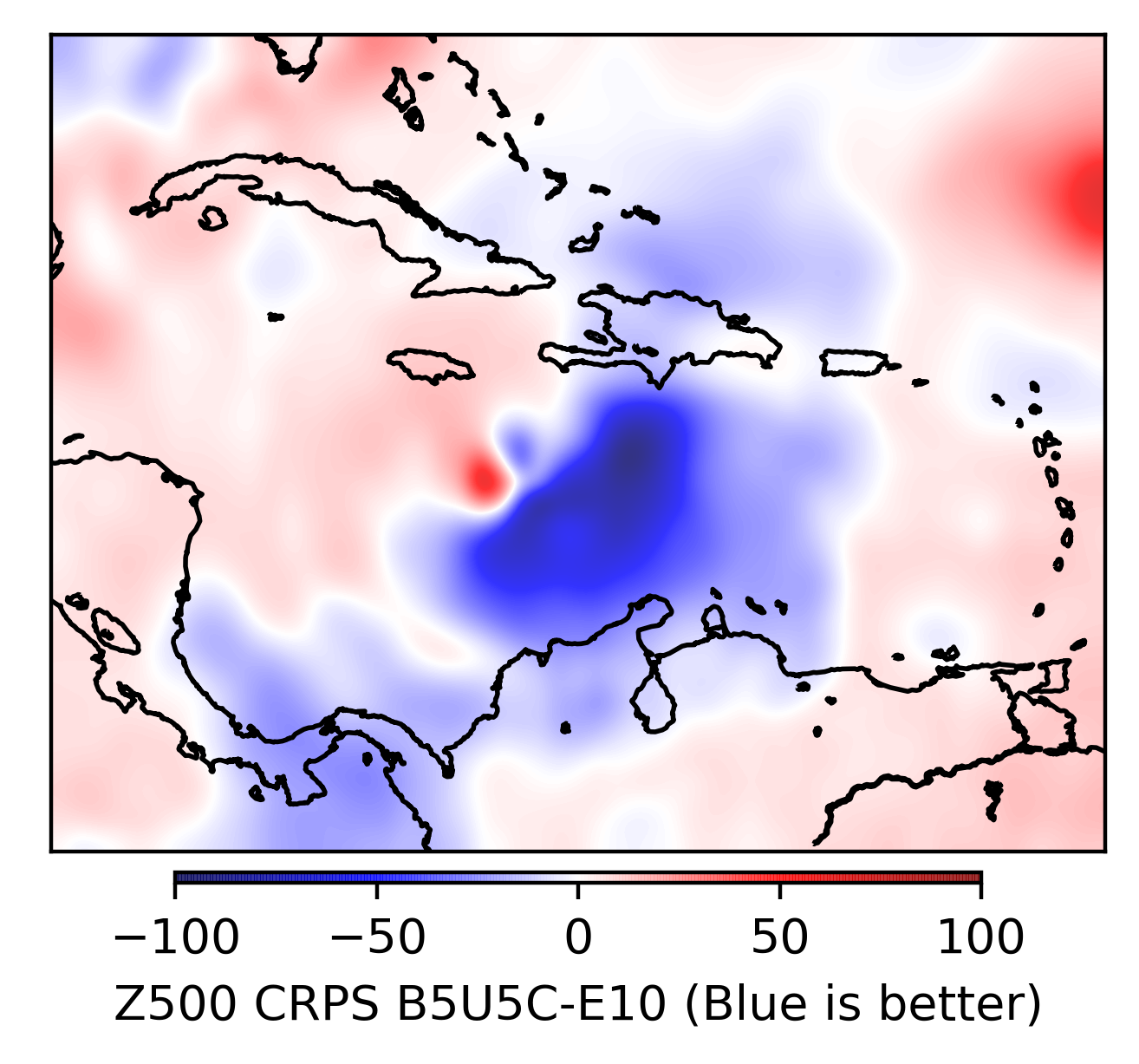}
    \caption{B5U5C-E10}
\end{subfigure}
\caption{Hurricane Matthew, a category five hurricane brought severe destruction to the Carribean and southeastern United States during September and October of 2016. We look at the Z500 forecast for the third of October. Again we see large improvements over the centre of the hurricane, as well as in the northern regions the hurricane will later progress over, with slightly reduced skill for outer regions.}
\label{fig:Matthew}
\vspace{-1em}
\end{figure*}

\begin{figure*}[h]
\centering
\begin{subfigure}{0.49\textwidth}
    \centering
    \includegraphics[width=1\linewidth]{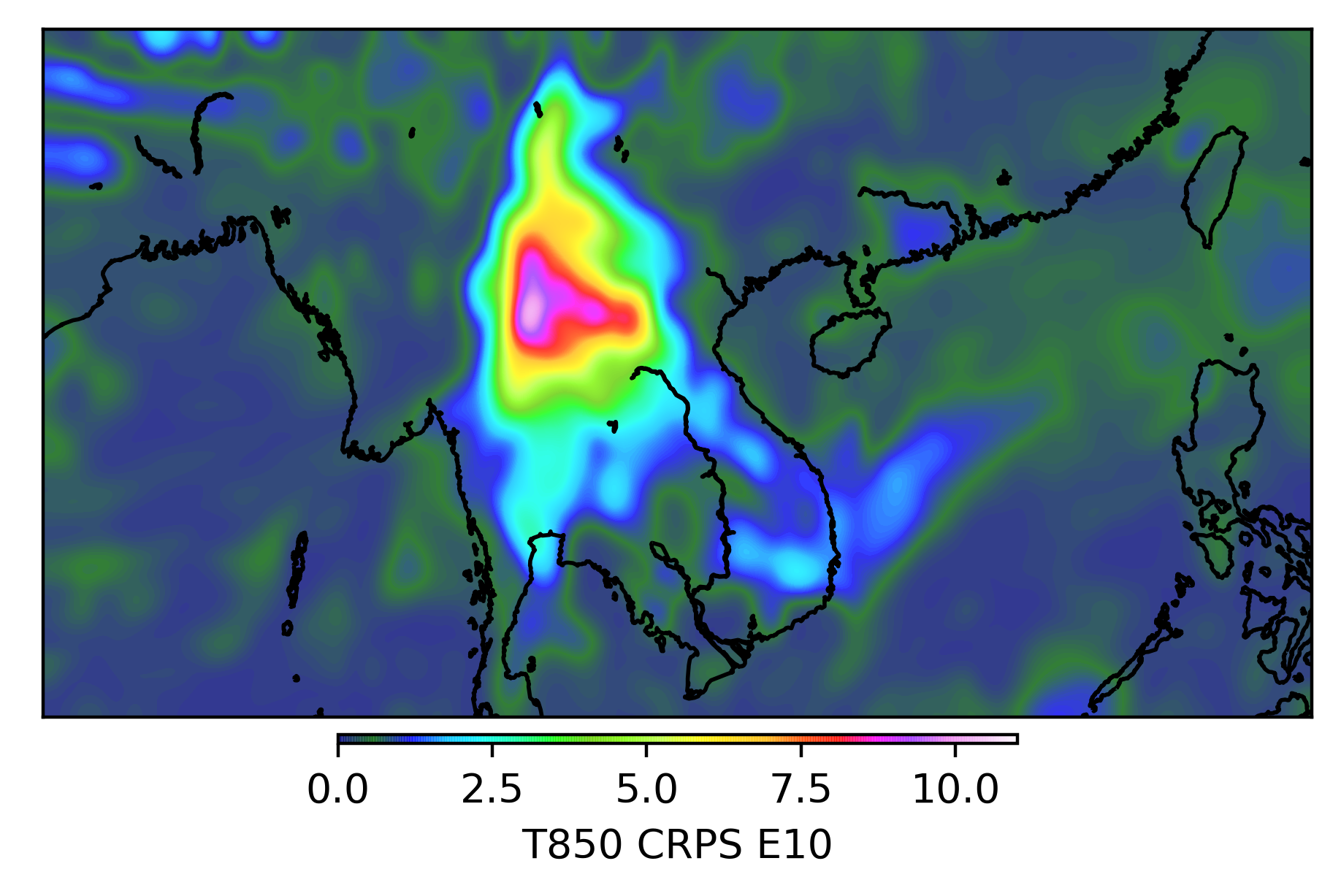}
    \caption{E10}
\end{subfigure}
\begin{subfigure}{0.49\textwidth}
    \centering
    \includegraphics[width=1\linewidth]{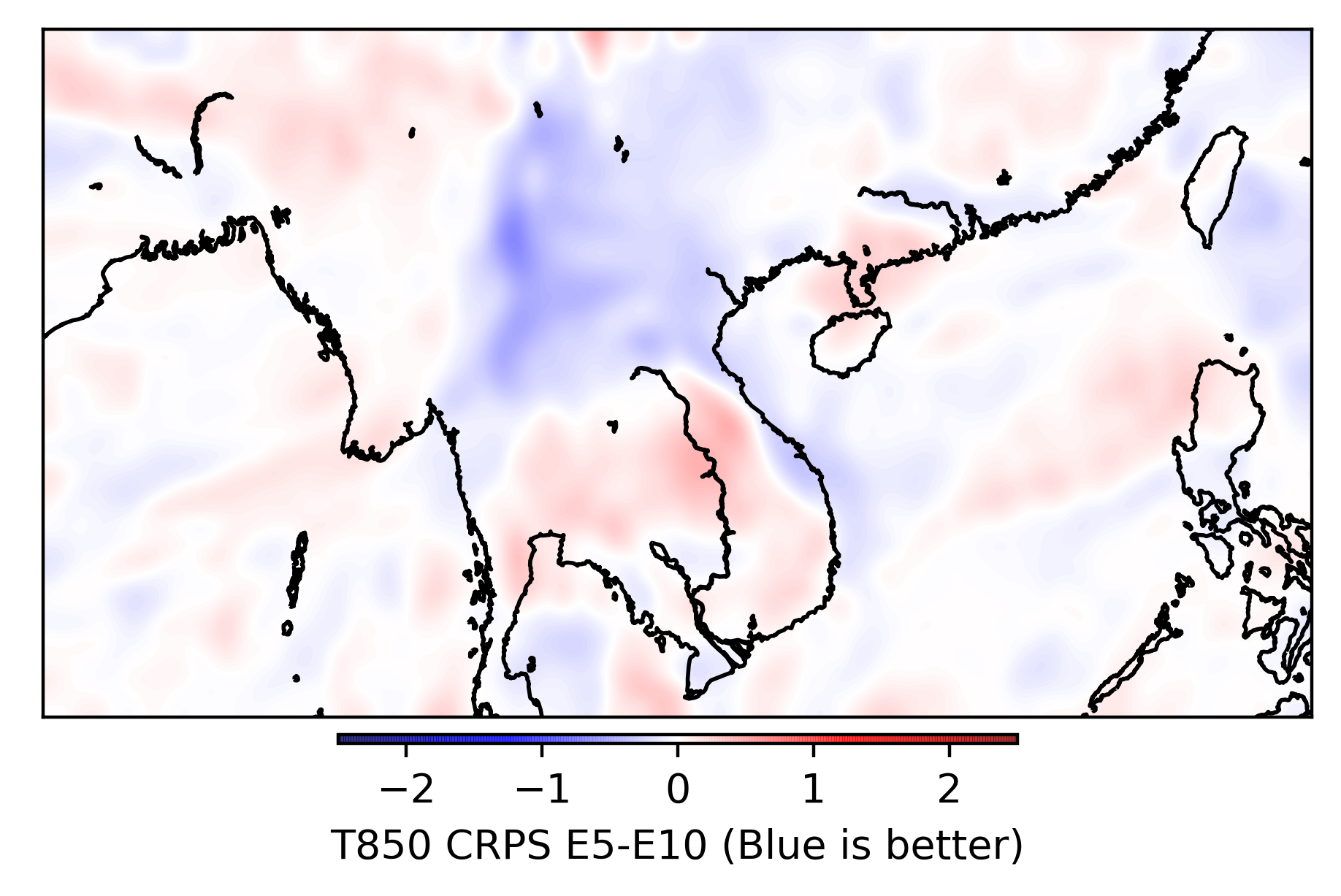}
    \caption{E5-E10}
\end{subfigure}\\
\vspace{-1em}
\begin{subfigure}{0.49\textwidth}
    \centering
    \includegraphics[width=1\linewidth]{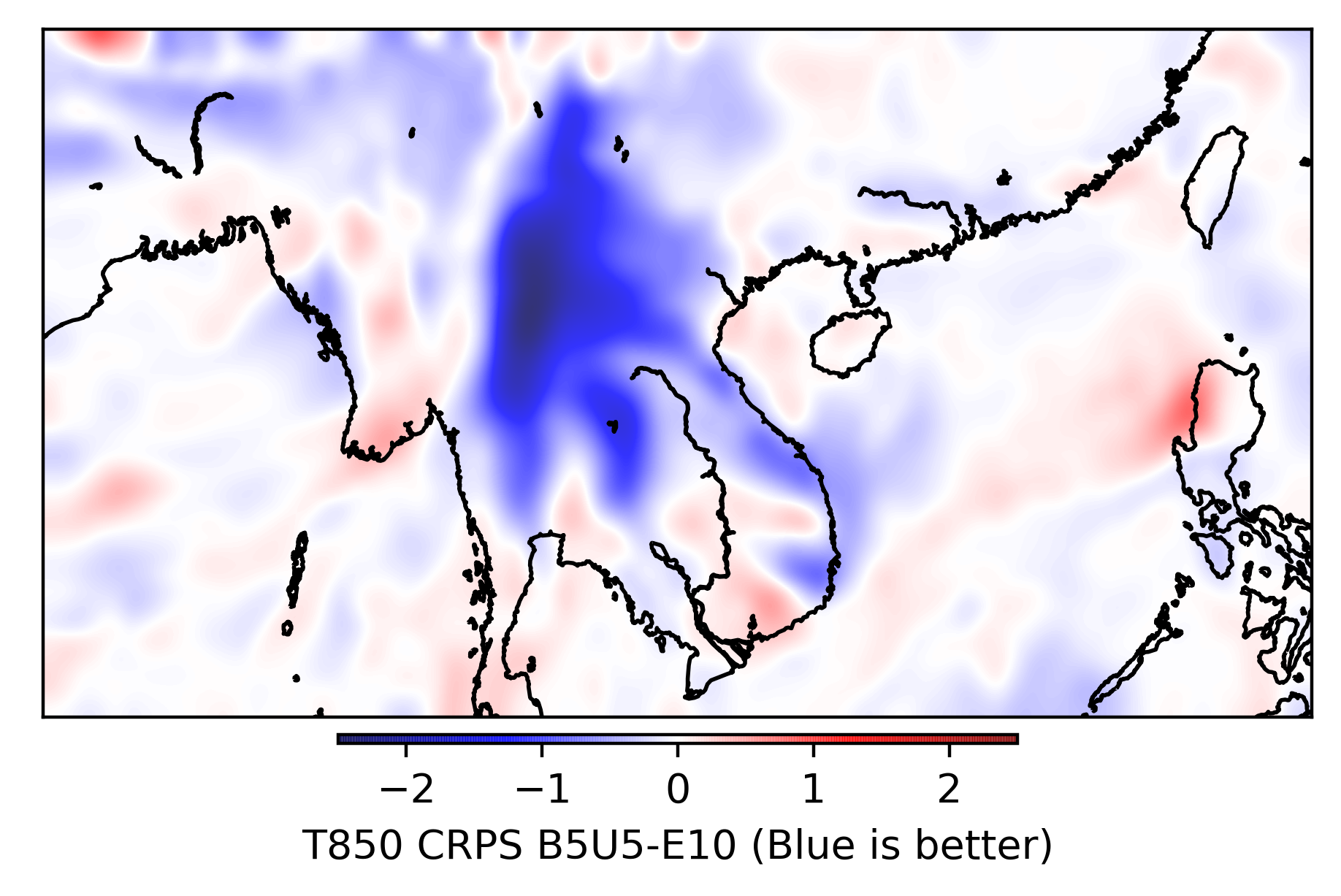}
    \caption{B5U5-E10}
\end{subfigure}
\begin{subfigure}{0.49\textwidth}
    \centering
    \includegraphics[width=1\linewidth]{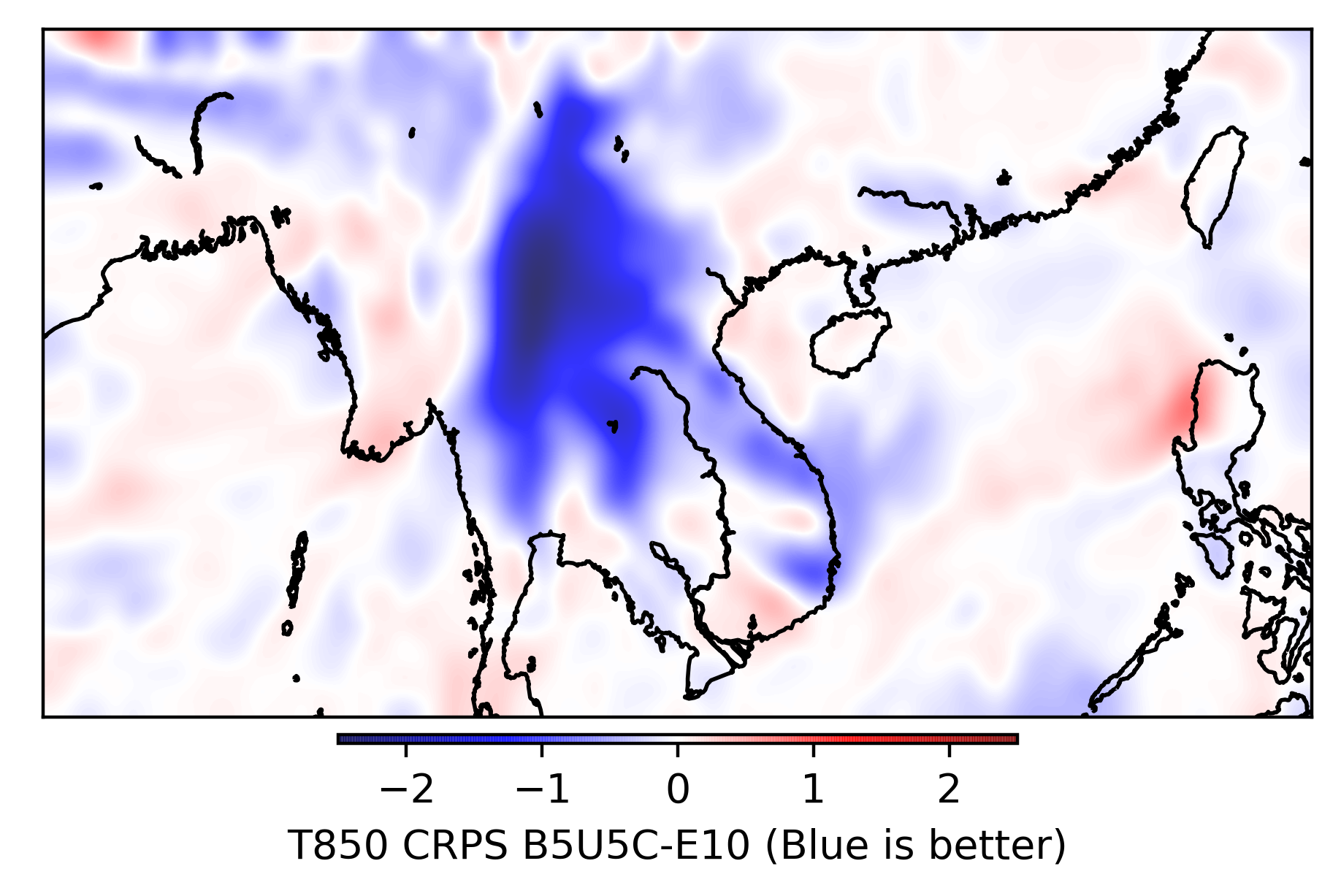}
    \caption{B5U5C-E10}
\end{subfigure}
\caption{Cold Wave over Asia: During January 2016 an unprecedented cold wave rushed over East and South Asia, leading to record lows. We focus on a forecast for January 24 where T850 forecast CRPS has its worst spike. In this case our CRPS trained model brings a large improvement of more than 25.5\% (CRPSS), compared to the 10 member ensemble, over the most affected region, while keeping regions of low CRPS fairly close to their original values, resulting in a total forecast improvement of around 19.5\% for the selected region.}
\vspace{-1em}
\label{fig:CFEA}
\end{figure*}

\begin{figure*}[h]
\centering
\begin{subfigure}{0.49\textwidth}
    \centering
    \includegraphics[width=1\linewidth]{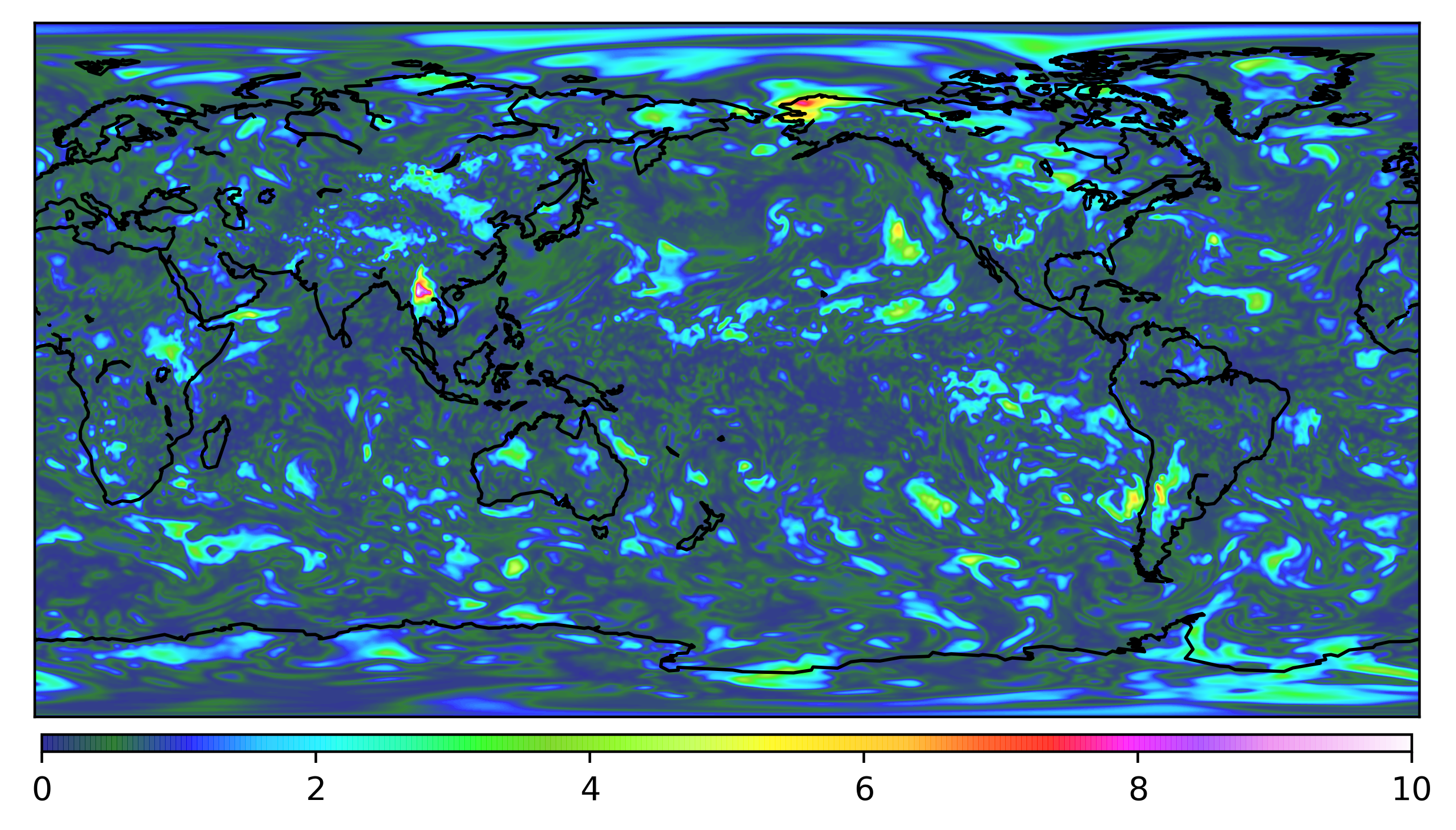}
    \caption{E10}
\end{subfigure}
\begin{subfigure}{0.49\textwidth}
    \centering
    \includegraphics[width=1\linewidth]{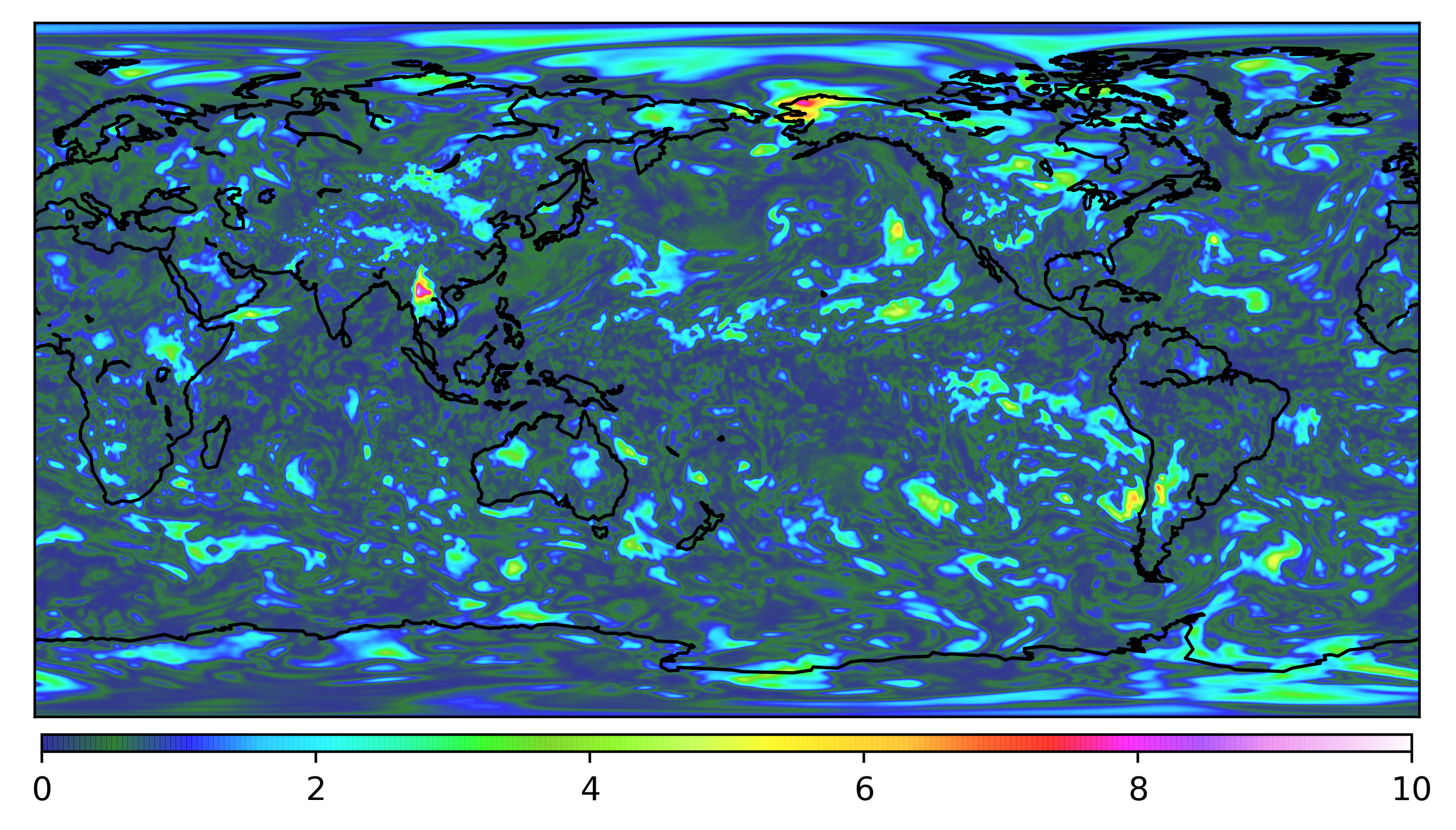}
    \caption{E5}
\end{subfigure}\\
\begin{subfigure}{0.49\textwidth}
    \centering
    \includegraphics[width=1\linewidth]{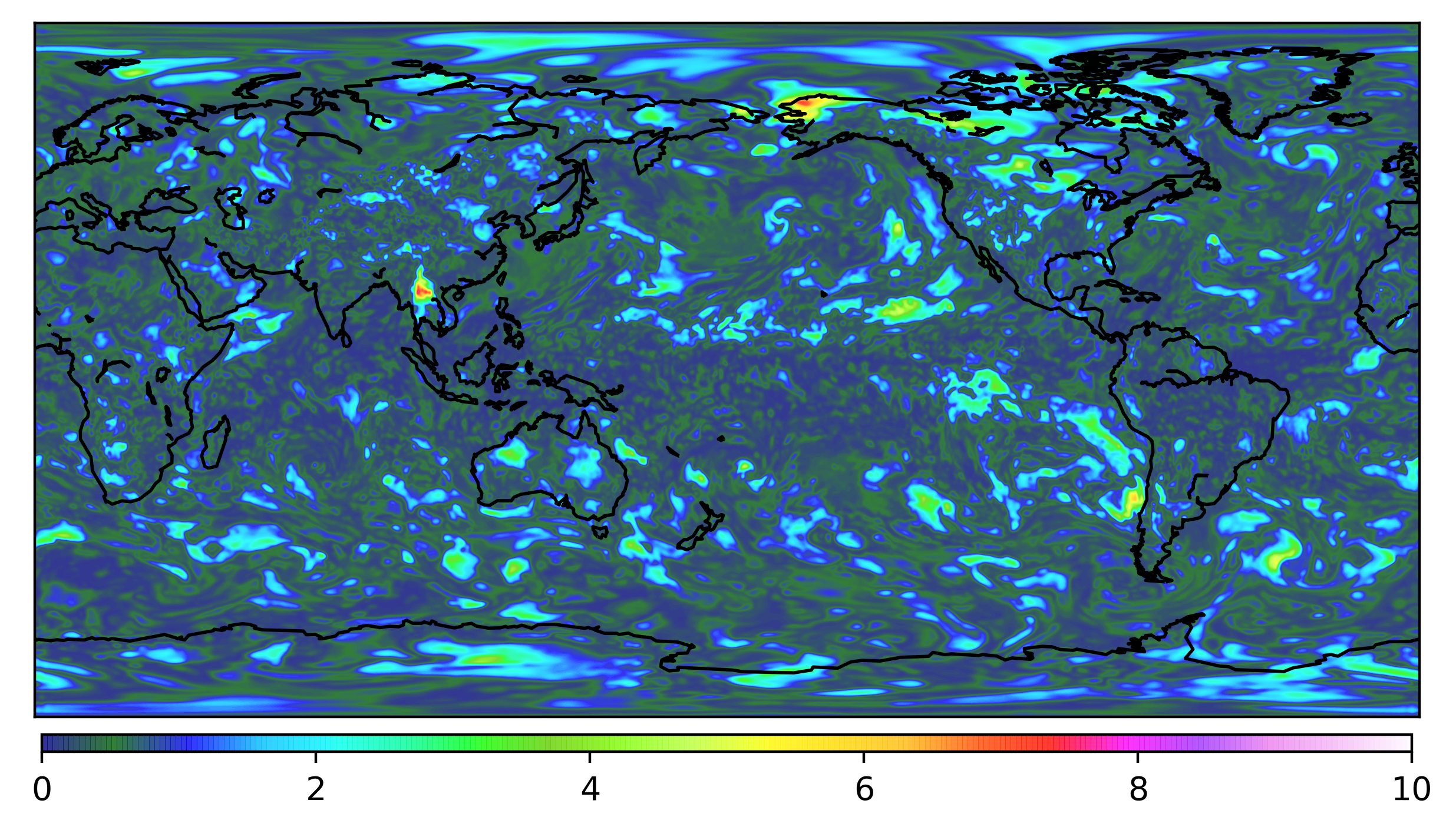}
    \caption{B5U5}
\end{subfigure}
\begin{subfigure}{0.49\textwidth}
    \centering
    \includegraphics[width=1\linewidth]{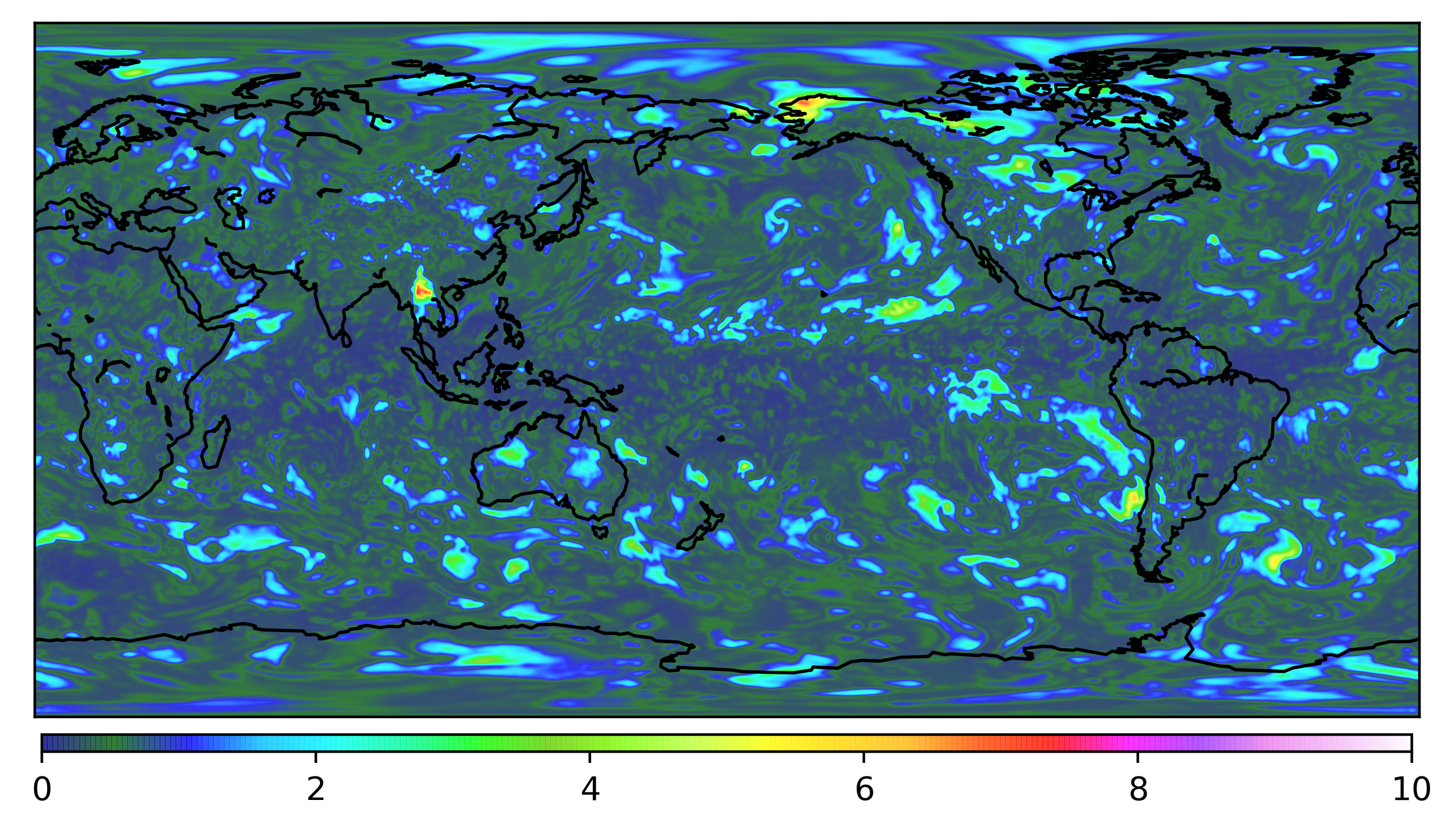}
    \caption{B5U5C}
\end{subfigure}
\caption{Global T850 CRPS plots for our models and the ENS10 forecast for January 24 2016 during the cold wave over East and South Asia (lower values are better).}
\label{fig:GLCR}
\end{figure*}

Thus far, our results have only demonstrated improvements on average values. They do not cover the performance for uncertainty quantification of extreme events, where forecast reliability is essential. In the following, we present three cases of extreme weather phenomena within our test set, selected from across the world, to demonstrate the networks' improvements for specific predictions. However, we would like to warn the reader that the interpretation of the probabilistic scores for specific events is difficult, as improvement or degradation for a single event are not sufficient to conclude that a method is better or worse in general. This limitation is visible in the following examples, as the E5 ensemble is able to outperform the E10 ensemble for specific events and locations. This can be explained by the limited number of ensemble members that have been used, and insufficient sampling of the probability distribution. Furthermore, the interpretation of predictions of extreme events by the end-user needs to be considered as well (see for example Lerch et al.~\cite{Lerch2017}). \Cref{fig:Winston,fig:Matthew,fig:CFEA} show our CRPS improvements for tropical cyclone Winston, hurricane Matthew, and a cold wave over Asia, respectively. Blue colour indicates that the (post-processed) five member ensemble has more skill when compared to the ten member ensemble. All times are 00:00 UTC. We also present global CRPS plots in Figure~\ref{fig:GLCR}.

\section{Conclusion}

We show that using informed model construction, deep learning can indeed improve the skill of global ensemble weather predictions. 
In particular, since we do not only predict an ensemble spread, but also perform a locally-adaptive output bias correction, we improve the results of a five-member ensemble to even surpass the forecast skill of a ten-member ensemble in terms of CRPS. 
When tasked with hard-to-predict extreme weather cases, such as tropical cyclone Winston, the combined models exhibit especially pronounced forecast skill improvements.  
Through the use of heterogeneous hardware, they are able to run these global post-processing steps within tenths of a second. 
In the future, such deep learning tools could allow for reduced ensembles to be run at higher resolutions, providing cheaper and more informed predictions.

The network structures used in this paper 
should also be tested for other applications of deep learning in NWP, such as the learning of model error in data-assimilation systems or the learning of the global equations of motion. 
Future research could be conducted into whether the networks need to be re-trained to process other physical fields and forecast lead times or whether the normalisation of spread values could allow the same network to also be applied to these tasks through transfer learning. 
Recurrent networks that encompass more time steps, as well as deep learning models that are capable of working on the native unstructured grid of the prediction model (e.g., graph neural networks~\cite{4700287}), can also be investigated in this context. 
Furthermore, the presented improvements would need to be studied when applied to ensembles with more members, such as the operational 50-member ensemble system of the ECMWF.
Lastly, while the fields that are investigated in this paper (Z500 and T850) are important to explore the potential capabilities of deep learning for this study, other fields that have more local dependence and output three-dimensional representations of the atmosphere, should be investigated in future work.

We encourage researchers to make use of the ERA5 and ENS10 datasets as well as our code, to apply new deep learning methods and expand on our initial architectures, helping weather forecast centres worldwide.

\section*{Acknowledgements}

We thank the Swiss National Supercomputing Centre for
providing compute resources and technical support.
Tal Ben-Nun is supported by the Swiss
National Science Foundation (Ambizione Project No. 185778).
Nikoli Dryden is supported by the ETH Postdoctoral Fellowship. This project also received funding from the European Research Council (ERC) under the
European Union’s Horizon 2020 programme (grant agreement DAPP, No. 678880 and EPiGRAM-HS, No. 801039).
Peter Dueben gratefully acknowledges funding from the Royal Society for his University Research Fellowship and the ESIWACE2 project. The ESIWACE2 has received funding from the European Union’s Horizon 2020 research and innovation programme under grant agreement No 823988.
%

\newpage
\appendix
\section{CRPS derivation}
\label{sec:CRPS_Equation}
The Cumulative Distribution Function (CDF) of a Normal Distribution can be written as: 
\begin{align*}
    F(x) &= \frac{1}{2}\left(1+\Phi\left(\frac{x-\mu}{\sigma\sqrt{2}}\right)\right)\\
    \Phi(x) &= \frac{2}{\sqrt{\pi}}\int_{0}^{x}e^{-t^2}dt\,,
\end{align*}
with $\Phi$ being the error function. The CRPS is then defined as:
\begin{align}
\text{CRPS}(F,y)=\int_{-\infty}^{\infty}[F(x)-\textbf{1}_{x>y}]^{2}dx\,,
\end{align}
which can be written as:
\begin{align}
&\int^{y}_{-\infty}{\left(\frac{1}{2}\left(1+\Phi\left(\frac{x-\mu}{\sigma\sqrt{2}}\right)\right)\right)^{2}}dx \\&+ \int^{\infty}_{y}{\left(\frac{1}{2}\left(1+\Phi\left(\frac{x-\mu}{\sigma\sqrt{2}}\right)\right)-1\right)^{2}}dx\,.
\end{align}
Continuing, we get:
\begin{align}
\begin{split}
&\left[\frac{1}{2}\left(\left(\sqrt{\frac{2}{\pi}}\sigma e^{-\frac{\left(\mu-x\right)^{2}}{2\sigma^{2}}} + x - \mu\right) \Phi\left(\frac{x-\mu}{\sqrt{2}\sigma}\right) + \frac{1}{2}\left(x-\mu\right) \Phi\left(\frac{x-\mu}{\sqrt{2}\sigma}\right)^{2} \vphantom{\frac{\sigma\Phi\left(\frac{\mu-x}{\sigma}\right)}{\sqrt{\pi}}}\right. \right. \\
&\left. \left. + \frac{\sigma\Phi\left(\frac{\mu-x}{\sigma}\right)}{\sqrt{\pi}} + \sqrt{\frac{2}{\pi}}\sigma e^{-\frac{\left(\mu-x\right)^{2}}{2\sigma^{2}}}+\frac{x}{2}\right)\right]^{y}_{-\infty}
\end{split}
\end{align}
from (A 2) and
\begin{align}
\begin{split}
&\left[\frac{1}{2}\left(\left(\sqrt{\frac{2}{\pi}}\sigma e^{-\frac{\left(\mu-x\right)^{2}}{2\sigma^{2}}} + \mu - x\right) \Phi\left(\frac{x-\mu}{\sqrt{2}\sigma}\right) + \frac{1}{2}\left(x-\mu\right)\Phi\left(\frac{x-\mu}{\sqrt{2}\sigma}\right)^{2} \vphantom{\frac{\sigma\Phi\left(\frac{\mu-x}{\sigma}\right)}{\sqrt{\pi}}}\right. \right. \\
& \left. \left. + \frac{\sigma\Phi\left(\frac{\mu-x}{\sigma}\right)}{\sqrt{\pi}} - \sqrt{\frac{2}{\pi}}\sigma e^{-\frac{\left(\mu-x\right)^{2}}{2\sigma^{2}}}+\frac{x}{2}\right)\right]^{\infty}_{y}
\end{split}
\end{align}
from (A 3). If we now place in the bounds and sum (A 4) and (A 5) up we arrive at:
\begin{align}
    \left(y-\mu\right)\Phi\left(\frac{y-\mu}{\sqrt{2}\sigma}\right)+\sqrt{\frac{2}{\pi}}\sigma e^{-\frac{\left(\mu-x\right)^{2}}{2\sigma^{2}}}
\end{align}
from inserting the $y$ bounds, and
\begin{align}
    -\frac{\sigma}{\sqrt{\pi}}
\end{align}
from inserting the $\infty$ bounds, resulting in:
\begin{align}
    \Delta{P} &= y - \mu = \text{Ground\_Truth} - \text{Prediction} \\
\text{CRPS}(\sigma,\Delta{P})&=\Delta{P}\cdot\Phi\left(\frac{\Delta{P}}{\sqrt{2}\sigma}\right) + \frac{\sigma}{\sqrt{\pi}}(-1+\sqrt{2}e^{-\frac{\Delta{P^2}}{2\sigma{^2}}})\,.
\end{align}

\end{document}